%% file: arxiv.tex
\documentclass[letterpaper]{article}
\usepackage[preprint]{aaai2027}
\usepackage[hyphens]{url}
\usepackage{graphicx}
\usepackage{natbib}
\usepackage{caption}
\usepackage{amsmath}
\usepackage{amssymb}
\usepackage{booktabs}
\usepackage{tabularx}
\usepackage{array}
\usepackage{dblfloatfix}

\title{From Faulty Memories to Corrected Actions: Dependency-Guided Rollback Repair for Memory-Augmented Agents}

\author{
{\small
Yiqi Wang\textsuperscript{*},
Caili Yu,
Jiaqi Zhang,
Yiqun Duan,
Mingkai Zheng,
Zhangkai Wu,
Kaize Shi,
Taotao Cai
}
\\[2pt]
{\footnotesize
\textsuperscript{*} yiqi.wang.jennie@gmail.com
}
}

\affiliations{}

\begin{document}

\maketitle

\begin{abstract}
Persistent memory lets language model agents reuse information across sessions, but it also makes errors durable: a poisoned, stale, or misattributed record can alter reasoning, tool use, answers, and subsequent memory writes.
Existing defenses mainly detect or delete suspicious memories, or revise the current response. Deleting the source leaves already propagated claims, actions, and derived memories active, whereas resetting the store or replaying the full trace destroys benign state and repeats unnecessary computation.
We therefore formulate \textbf{post-failure memory recovery}: \textit{given a failed execution and diagnosed faulty memories, regenerate the answer and clean up affected persistent state while retaining unaffected work.}
Our \textbf{dependency-guided rollback repair} builds a typed memory-to-action graph from runtime provenance, traces explicit downstream dependencies, preserves candidates with independent trusted support, deactivates unsupported memory state, and selectively replays only answer-relevant affected computation.
We evaluate this approach on a 150-case controlled benchmark spanning three tool-use domains and four memory failure types, and on a 50-case trajectory-derived stress test adapted from LongMemEval-V2.
On the controlled benchmark, it achieves 85.3\% immediate recovery versus 77.3\% for LLM-judge repair, the strongest baseline, while preserving all benign memories and using 41.8\% fewer LLM calls. Selective replay, however, leaves higher recurrence than LLM-judge repair (26.6\% vs.\ 12.1\%), exposing a trade-off between targeted repair and broader recomputation.
On the adapted subset, it reaches 68.0\% recovery versus 54.0\% for the next best method, with 5.9\% recurrence.
Code and data are available at \url{https://github.com/xiaoqi-7/memory-rollback}.
\end{abstract}

\input{sections/01_intro}
\input{sections/02_related_work}
\input{sections/03_method}
\input{sections/04_benchmark}
\input{sections/05_experiments}
\input{sections/06_conclusion}

\bibliography{references}

\clearpage
\input{appendix}

\end{document}

%% file: sections/01_intro.tex
\section{Introduction}
\label{sec:introduction}

Persistent memory lets language model agents carry preferences, observations, and experience across sessions, enabling personalization and long-horizon behavior that a current prompt alone cannot support~\cite{park2023generative,packer2023memgpt,zhang2025survey,zhong2024memorybank,wang2023voyager}. The same persistence changes the scope of failure. A poisoned, stale, misattributed, or drifted record can be retrieved as context, support an incorrect claim, alter a tool plan and answer, and be consolidated into new memory~\cite{xiong2026memory,chao2026stale,chen2025halumem}. A single local fault can therefore become a durable family of downstream errors and reappear after the original turn.

\begin{figure}[t]
\centering
\includegraphics[width=\columnwidth,keepaspectratio]{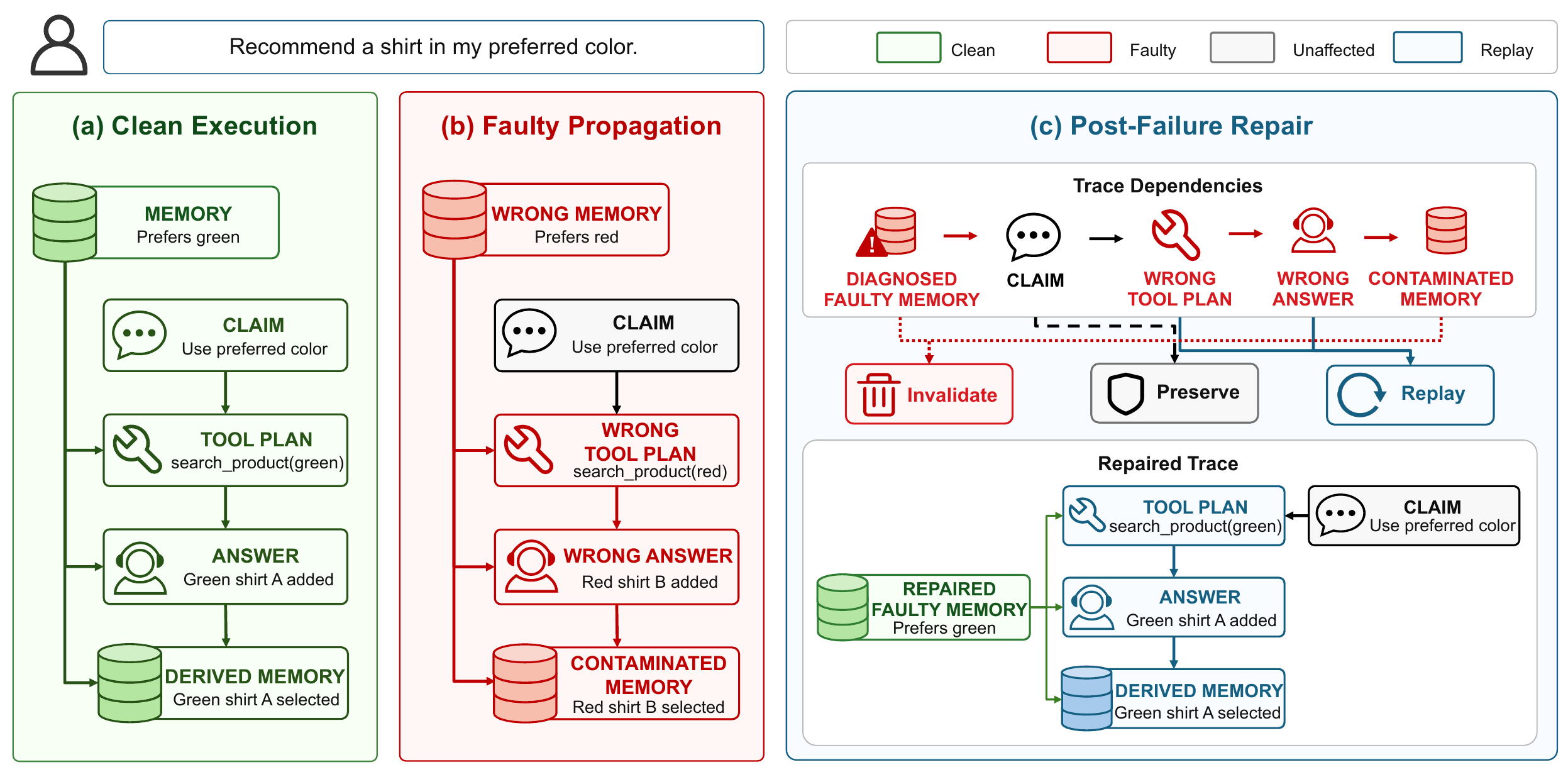}
\caption{\textbf{A faulty memory creates downstream state.} A poisoned preference changes the claim, tool plan, answer, and a derived memory. Repair must remove unsupported descendants, preserve unaffected state, and replay the answer-relevant path.}
\label{fig:problem}
\end{figure}

This propagation creates the practical recovery problem illustrated in Figure~\ref{fig:problem}. Correcting only the answer leaves the faulty source and contaminated descendants available to influence future turns. Deleting only the source is also insufficient after its content has been copied into claims, observations, or derived memories. At the other extreme, clearing the memory store or replaying the full trace removes valid personalization and repeats model and tool calls. Useful recovery must therefore remove unsupported consequences, retain state that still has independent evidence, and recompute only what the corrected answer requires.

Existing research largely acts at one of two boundaries. Memory defenses audit, filter, or delete suspicious records, while response-level reflection revises an unsuccessful output~\cite{tan2026memaudit,zou2025poisonedrag,shinn2023reflexion}. These interventions improve detection or local correction, but do not jointly repair the persistent store and the execution state already influenced by a fault. This gap motivates our central premise: \emph{fault removal is not state recovery}.

We therefore formulate \textbf{post-failure memory recovery}.
The input is a user session, an active memory store, a failed execution trace, and diagnosed faulty memories supplied by an upstream detector;
the output is a regenerated answer, repaired trace, and repaired active store. Diagnosis itself is outside the scope of this work, and the runtime must record the provenance connecting memory reads, claims, plans, actions, observations, answers, and mutations. Under these assumptions, the research question is precise: \emph{once a faulty memory has been consumed, which consequences should be invalidated, which remain independently justified, and which must be regenerated?}

We propose a \textbf{dependency-guided rollback repair}. The method first constructs a typed memory-to-action graph and traces explicit downstream dataflow from each diagnosed fault. Because reachability alone would over-invalidate, it preserves candidates whose complete content or decision has independent trusted support. A deterministic planner then deletes diagnosed faults, quarantines unsupported derived state, invalidates unsupported trace outputs, and selects the answer-relevant affected computation. Finally, the executor reuses safe context and replays selected steps in trace order under the repaired store. The design directly couples the three requirements above: contamination tracing, evidence-aware preservation, and selective recomputation.

We evaluate answer recovery, state repair, preservation, and cost on 150 controlled cases across shopping, travel, and customer-support tools, covering four memory fault types, and on 50 procedural trajectories adapted from LongMemEval-V2 as a multi-fault stress test~\cite{wu2026longmemeval}. We distinguish three levels of state repair: removing the diagnosed source faults, deactivating identified unsupported downstream state, and preventing the failure from recurring when the repaired store is reused. Our method targets the first two together with \emph{immediate recovery}, i.e., a correct regenerated answer for the failed turn, and we measure residual recurrence separately. It obtains the highest immediate recovery in both settings with lower cost than LLM-judge repair, but on the controlled benchmark its recurrence (26.6\%) is higher than that of LLM-judge repair (12.1\%), reflecting a trade-off between targeted and broader recomputation.

Our main contributions are summarized as follows:
\begin{itemize}
    \item We formulate \textbf{post-failure memory recovery}, which jointly considers answer recovery, downstream state repair, benign-state preservation, and repair cost after faulty memories have already affected agent execution.

    \item We propose a \textbf{dependency-guided rollback repair} method that, beyond deactivating faulty and derived memories, determines which downstream claims and tool computations remain independently supported and which affected executable prerequisites must be replayed to regenerate the final answer.

    \item We introduce a \textbf{150-case controlled benchmark} plus a 50-case trajectory-derived stress test, showing the highest immediate recovery and a favorable recovery--cost trade-off, at the price of higher recurrence than broader LLM-judge repair on the controlled benchmark.

\end{itemize}

%% file: sections/02_related_work.tex
\section{Related Work}

\subsubsection{Agent memory and failure detection.} Persistent-memory systems support cross-session personalization and experience reuse~\cite{park2023generative,packer2023memgpt,xu2026mem,du2026memguide,zhao2026ama}. Recent work studies error propagation, hallucinated updates, stale state, retrieval or memory poisoning, and post-hoc auditing~\cite{xiong2026memory,chen2025halumem,chao2026stale,zou2025poisonedrag,chen2024agentpoison,tan2026memaudit}. These methods primarily improve what enters, remains in, or is retrieved from memory. We instead condition on diagnosed faults that have already influenced an execution.

\subsubsection{Memory cascade repair.}
MemoRepair is the closest memory-side recovery framework: it withdraws invalidated descendants, rebuilds successors from retained support, and uses predecessor closure for cost-aware republication~\cite{zhao2026memorepair}. Once a fault has crossed into agent execution, two further decisions arise that memory-side repair does not address. Beyond withdrawing derived memories, our method determines which claims, plans, and tool computations remain independently supported, and which affected executable prerequisites, including tool actions whose observations must be refreshed, must be replayed to regenerate the final answer. MemoRepair's publication contract and fault detector are orthogonal to this setting.

\subsubsection{Provenance, slicing, and rollback.}
Forward dependency tracing and selective recomputation draw on program and dynamic slicing~\cite{weiser1984program,agrawal1990dynamic}, database provenance~\cite{cheney2009provenance}, and transactional rollback~\cite{mohan1992aries}. We do not claim graph reachability or logical undo as new. What differs in our setting is that a reachable node may remain valid through independent evidentiary support, and that each replayed step incurs model and tool cost, so tracing must be combined with support checking and answer-relevant replay selection.

\subsubsection{Trace debugging and response repair.}
ReAct exposes reasoning and tool-use structure, while Reflexion and Self-Refine revise unsuccessful behavior or outputs~\cite{yao2022react,shinn2023reflexion,madaan2023self}. AgentTrace reconstructs causal graphs from logs for root-cause localization~\cite{wang2026agenttrace}. These ideas motivate trace-aware recovery, but localization or answer revision alone need not deactivate contaminated persistent state. Our dependency graph therefore spans the memory lifecycle and execution trace, and its rollback plan couples memory disposition with selective replay.

%% file: sections/03_method.tex
\section{Dependency-Guided Rollback Repair}
\label{sec:method}

\begin{figure*}[t]
\centering
\includegraphics[width=0.9\textwidth]{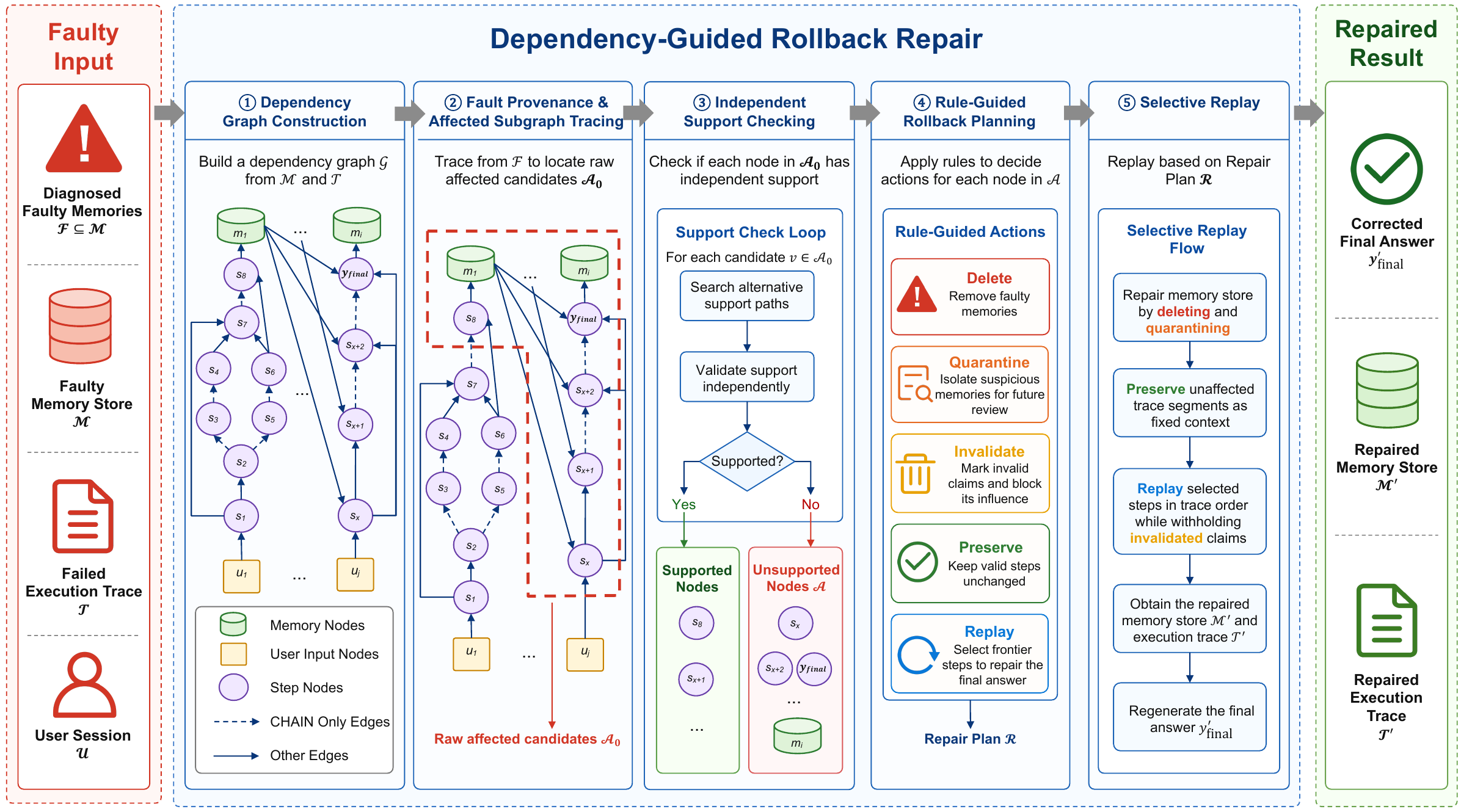}
\caption{\textbf{Overview of dependency-guided rollback repair.} Starting from a diagnosed faulty memory set, the method constructs a memory-to-action dependency graph, traces candidate downstream effects, filters candidates with independent trusted support, generates a rule-guided rollback plan, and selectively replays answer-relevant affected computation to produce a corrected final answer and a repaired active memory store.}
\label{fig:method-overview}
\end{figure*}

As Figure~\ref{fig:method-overview} shows, dependency-guided rollback repair proceeds in five stages: graph construction, affected-subgraph tracing, independent-support checking, rule-guided rollback planning, and selective replay. Rollback here means repair of agent-maintained trace and memory state. It cannot undo an irreversible external side effect. Re-executing a side-effecting tool therefore requires a resettable interface or a domain-specific compensating action. Full schemas are in Appendix~\ref{app:data-schema} and replay prompt contracts in Appendix~\ref{app:prompt}.

\subsection{Problem Setting}
\label{sec:problem-definition}

Let $\mathcal{U}=(u_1,\ldots,u_k)$ be a user session, $\mathcal{M}$ the active memory store, and $\mathcal{T}=(s_1,\ldots,s_n)$ a failed execution trace. A step is a memory read, claim, plan, tool action, tool observation, answer, or memory mutation. The input also contains diagnosed faulty memories $\mathcal{F}\subseteq\mathcal{M}$. At repair time, our method sees only $(\mathcal{U},\mathcal{M},\mathcal{T},\mathcal{F})$, available tools, and runtime provenance; clean traces and evaluation labels are withheld. The method outputs a repair plan $\mathcal{R}=(\mathcal{M}_{\mathrm{del}},\mathcal{M}_{\mathrm{quar}},\mathcal{V}_{\mathrm{inv}},\mathcal{V}_{\mathrm{replay}},\mathcal{V}_{\mathrm{preserve}})$, specifying memories to delete or quarantine and trace steps to invalidate, replay, or preserve as fixed context (Section~\ref{sec:rollback-planning}), and executes it to produce a regenerated answer, a repaired trace, and a repaired active store. Quarantined memories are kept for auditing but excluded from retrieval and execution.

\subsection{Dependency Graph Construction}
\label{sec:dependency-graph}

We instantiate a directed heterogeneous graph $\mathcal{G}=(\mathcal{V},\mathcal{E})$ from user inputs $\mathcal{U}$, execution steps $\mathcal{T}$, and memory records $\mathcal{M}$. Construction is fault-agnostic: it records execution dataflow and memory-lifecycle structure without using $\mathcal{F}$ or any benchmark annotation. The node set is partitioned as $\mathcal{V}=\mathcal{V}_{\mathcal{U}}\cup\mathcal{V}_{\mathcal{T}}\cup\mathcal{V}_{\mathcal{M}}$, containing user inputs, typed execution steps, and persistent memory records; we identify each memory record with its node in $\mathcal{V}_{\mathcal{M}}$ when using memory sets in graph expressions.

We instantiate edges deterministically from these fields rather than inferring them. Edges point from a prerequisite or source to its dependent. They encode \textsc{initiate}, \textsc{cite}, \textsc{support}, \textsc{produce}, \textsc{delete}, \textsc{update}, \textsc{consolidate} relations; and memory \textsc{supersede}, \textsc{derive} relations. Temporal \textsc{chain} edges retain execution order. A node pair may carry multiple labels. Appendix~\ref{app:data-schema} gives the full edge semantics.

Each trace record stores the identifiers of the memories, claims, observations, together with tool-call identifiers and memory-lifecycle metadata. This provenance is emitted by the agent runtime, so the method presupposes an instrumented agent. Inferring missing support or lifecycle edges from raw natural-language logs is outside the scope of this work; incomplete provenance can under-trace contamination, while spurious edges can cause unnecessary invalidation.

For propagation we use
\[
\begin{aligned}
\mathcal{E}_{\mathrm{prop}}={}&
\mathcal{E}_{\mathrm{initiate}}\cup\mathcal{E}_{\mathrm{cite}}\cup \mathcal{E}_{\mathrm{support}}\cup\mathcal{E}_{\mathrm{produce}}\\
&{}\cup\mathcal{E}_{\mathrm{delete}}\cup\mathcal{E}_{\mathrm{update}}\cup
\mathcal{E}_{\mathrm{consolidate}}\\
&{}\cup\mathcal{E}_{\mathrm{supersede}}\cup\mathcal{E}_{\mathrm{derive}}.
\end{aligned}
\]
Pure \textsc{chain} edges are excluded: occurring later is not evidence of contamination. This distinction is important because a sequential suffix may contain unrelated valid work.

\subsection{Fault Provenance \& Affected Subgraph Tracing}
\label{sec:affected-tracing}

Given $\mathcal{F}$, we identify the state that may depend on these faults. This stage does not use fault-type labels such as poisoned, stale, wrong-user, or summary-drift; all diagnosed faults are handled uniformly through graph structure and memory provenance.

For each $m \in \mathcal{F}$, we use fault provenance to
identify a reachability seed $r(m)$. We inspect \textsc{delete}, \textsc{update}, and \textsc{consolidate} relations to recover lineage and identify mutations. If a
failed mutation exists, it becomes the seed; otherwise, we
follow a \textsc{produce} relation to the producing step. If $m$ was not created during the recorded execution and has no such producing step, the memory itself becomes the seed. When $r(m)$ occurs at time $t_r$ in the trace, only dependents timestamped no earlier than $t_r$ are considered.

We then trace the affected subgraph from each $r(m)$ and take the union of the reached nodes and all seeds to form the raw affected candidate set:
\[
\mathcal{A}_0
=
\bigcup_{m\in\mathcal{F}}
\mathrm{Reach}_{(\mathcal{V},\mathcal{E}_{\mathrm{prop}})}
\bigl(r(m)\bigr).
\]
A node is not marked as affected merely because it follows a faulty memory in execution order; it must be connected through an explicit propagation dependency. Favoring recall, $\mathcal{A}_0$ over-approximates the truly unsupported state and is refined next.

\subsection{Independent-Support Checking}
\label{sec:support-checking}

A node may be reachable from a faulty memory while remaining valid because it is also justified by independent, fault-free evidence. For example, a claim may cite a faulty memory together with an unaffected tool observation that supports the same conclusion. Treating every reachable node as contaminated would therefore cause unnecessary invalidation and replay.

For each checkable step $v\in\mathcal{A}_0$, namely a claim, plan, final answer, or memory mutation, we collect its direct evidentiary predecessors $\mathrm{Pred}_{\mathrm{sup}}(v)$: the memories connected to $v$ by \textsc{support} edges and the claims, tool observations, or user inputs connected by \textsc{cite} edges, i.e., the identifiers recorded in the step's provenance. A memory that $v$ itself deletes, updates, or consolidates is excluded, since a mutation target is not evidence for its own retirement. We define
\[
\begin{aligned}
\mathrm{TrustedSupport}(v)\iff{}&\exists\,p\in\mathrm{Pred}_{\mathrm{sup}}(v):\\
&p\notin\mathcal{F}\cup\mathcal{A}_0\ \land\ p\notin\mathcal{V}_{\mathcal{U}}.
\end{aligned}
\]
User inputs are excluded because they trigger a step rather than independently verify it. Admissibility is a policy assumption, not a guarantee of factual correctness: an active memory that has not been diagnosed as faulty is treated as admissible, not as verified. The check is structural and deliberately lenient. A single admissible predecessor suffices, and the check does not verify that this predecessor covers the whole content of $v$; for example, a plan fixing both a destination and a budget is preserved if it cites one unaffected source, even when that source supports only the destination. Memory trust scores are recorded but not used (Appendix~\ref{app:data-schema}).

Diagnosed faulty memories are never eligible for preservation. An active affected memory with \textsc{derive} or \textsc{produce} provenance is a derived-memory candidate; it is supported if a backward path through its producing step and \textsc{support}, \textsc{cite}, or \textsc{derive} edges reaches a trusted root outside $\mathcal{F}$, namely a user input, an unaffected tool observation, or an active root memory without derivation provenance. \textsc{supersede} edges are not followed, since replacing a memory is not evidence for it. Actions inherit the verdicts of their plans, and observations inherit those of their generating actions, so an independently supported plan may keep its unchanged action and observation. Only evidence outside $\mathcal{A}_0$ may preserve a step, and preserved candidates are not recursively reused to validate others, preventing mutually dependent affected nodes from validating one another. Writing $\mathcal{P}_{\mathrm{sup}}\subseteq\mathcal{A}_0$ for the preserved candidates, including associated safe action and observation nodes, the unsupported affected set is $\mathcal{A}=\mathcal{A}_0\setminus\mathcal{P}_{\mathrm{sup}}$: $\mathcal{P}_{\mathrm{sup}}$ stays eligible for reuse and $\mathcal{A}$ passes to the planner.

\subsection{Rule-Guided Rollback Planning}
\label{sec:rollback-planning}

The rollback planner converts the diagnosed faults, unsupported affected set, provenance information, and support verdicts into the repair plan $\mathcal{R}$. The planner is deterministic and rule-guided rather than optimization-based. Its memory dispositions and trace actions are orthogonal: deleting or quarantining a memory changes the active memory state, whereas invalidating, replaying, or preserving a trace node determines how the previous execution is treated.

The planner first deactivates unsafe persistent state, setting $\mathcal{M}_{\mathrm{del}}=\mathcal{F}$ and $\mathcal{M}_{\mathrm{quar}}\subseteq(\mathcal{A}\cap\mathcal{V}_{\mathcal{M}})\setminus\mathcal{F}$, so faulty memories leave the active store and support-preserved memories stay active. An unsupported derived memory is quarantined only if it still carries a fault-specific fact, i.e., a number, identifier, date, or status term, extracted by pattern matching, that appears in a diagnosed faulty memory but in no clean active memory or clean parent; otherwise it remains active. In addition, when a faulty memory has a recoverable producing step, or stale state results from a failed lifecycle operation, that step stays eligible for replay. If a support verdict conflicts with the provenance-seed rule, the latter takes precedence: the step that produced a diagnosed faulty memory is replayed even when its own inputs are independently supported, so that a corrected replacement is generated. All unsupported trace nodes are then marked invalid, $\mathcal{V}_{\mathrm{inv}}=\mathcal{A}\cap\mathcal{V}_{\mathcal{T}}$, so their outputs cannot serve as evidence; plans pass invalidation to their actions, and actions pass it to their observations, and observations are never replayed directly, only refreshed by scheduling the generating action. Appendix~\ref{app:rollback-rules} tabulates the full rule set.

To avoid replaying the full affected subgraph, the planner next identifies the answer-relevant affected computation. Let $v_{\mathrm{final}}$ denote the final answer of the last turn, the outcome being repaired, and let
\[
\mathcal{G}^{-}
=
\mathcal{G}
\left[
\mathcal{V}
\setminus
\left(
\mathcal{M}_{\mathrm{del}}
\cup
\mathcal{M}_{\mathrm{quar}}
\right)
\right]
\]
denote the subgraph of $\mathcal{G}$ induced after deleted and quarantined memories are removed as severed nodes. We define
\[
\mathcal{A}_{\mathrm{ans}}
=
\left\{
v\in
\mathcal{A}
\cap
\mathcal{V}_{\mathcal{T}}
\;\middle|\;
v\leadsto v_{\mathrm{final}}
\text{ in }
\mathcal{G}^{-}
\right\}.
\]
Unlike contamination tracing, this test also follows \textsc{chain} edges, so a step that precedes $v_{\mathrm{final}}$ in execution order is conservatively treated as able to influence it, even without explicit provenance. Affected final answers of earlier turns are left unchanged. Nodes outside $\mathcal{A}_{\mathrm{ans}}$ remain invalidated but are not replayed; the implementation records them as suspicious steps for auditing (Appendix~\ref{app:data-schema}).

The executable replay set is obtained by taking the execution closure of the answer-relevant affected nodes:
\[
\mathcal{V}_{\mathrm{replay}}
=
\mathrm{ExecClosure}
\left(
\mathcal{A}_{\mathrm{ans}}
\cup
\left\{
v_{\mathrm{final}}
\right\}
\right).
\]
The execution closure recursively adds every invalidated executable prerequisite required to recompute a selected node, including controlling plans, generating tool actions, producer steps, and memory mutation steps. Safe prerequisites are not replayed and are instead supplied through $\mathcal{V}_{\mathrm{preserve}}$. An affected memory read is replayed when the step it feeds is replayed. When a selected node is a tool observation, the execution closure adds its generating tool action rather than the observation itself. The final answer node is always included so that the repaired execution produces a fresh answer.

We define the preserved trace context as
\[
\mathcal{V}_{\mathrm{preserve}}
=
\mathcal{V}_{\mathcal{T}}
\setminus
\left(
\mathcal{V}_{\mathrm{inv}}
\cup
\mathcal{V}_{\mathrm{replay}}
\right).
\]
Therefore, $\mathcal{V}_{\mathrm{preserve}}$ contains safe prior execution outputs needed as fixed context during replay. User inputs and active memory records are supplied separately through $\mathcal{U}$ and the repaired memory store.

\subsection{Selective Replay}
\label{sec:selective-replay}

Given the repair plan $\mathcal{R}$, the selective replay executor first applies the planned memory dispositions. The active pre-replay memory store is
\[
\mathcal{M}^{-}
=
\mathcal{M}
\setminus
\left(
\mathcal{M}_{\mathrm{del}}
\cup
\mathcal{M}_{\mathrm{quar}}
\right).
\]
Records in $\mathcal{M}_{\mathrm{quar}}$ may be retained in a separate audit partition, but they are unavailable to retrieval and execution.

The executor processes $\mathcal{V}_{\mathrm{replay}}$ in the original trace order, thereby preserving the observed execution ordering. Memory-lifecycle relations used only for fault resolution are not treated as replay-precedence constraints. Nodes in $\mathcal{V}_{\mathrm{preserve}}$ are supplied as fixed context, while invalidated nodes outside the replay set remain excluded from subsequent execution.

Each selected step is recomputed using the current repaired memory state and the repaired prefix of the trace. The memory state is initialized as $\mathcal{M}^{-}$ and updated as replayed memory mutations are executed. Invalidated claims are withheld from the replay context until replacements are generated, preventing unsupported outputs from being cited by later steps. Replayed memory reads retrieve only from the current repaired store, and replayed memory mutations are applied only when they remain justified by the repaired execution.

Tool use is also replayed selectively. If the repaired plan still requires tool use, the executor issues a new action and records the resulting observation as a new trace node. Existing observations are never independently regenerated or treated as reusable when their generating actions have been invalidated. For side-effecting tools, re-execution is performed only through the resettable, idempotent, or compensating interface assumed in the problem setting.

Let $\Delta\mathcal{M}_{\mathrm{replay}}$ denote the ordered sequence of valid memory writes, deletions, updates, and consolidations generated during replay. The final repaired active memory store is
\[
\mathcal{M}'
=
\operatorname{Apply}
\left(
\mathcal{M}^{-},
\Delta\mathcal{M}_{\mathrm{replay}}
\right),
\]
where $\operatorname{Apply}$ executes the replayed memory mutations in trace order and removes records superseded by a valid delete, update or consolidation.

The final-answer node $v_{\mathrm{final}}$ is always regenerated from the repaired trace. Let $\mathcal{T}'$ denote the resulting execution trace and $y'_{\mathrm{final}}$ the content of the regenerated final answer node. The executor outputs the regenerated answer $y'_{\mathrm{final}}$, repaired execution trace $\mathcal{T}'$, and repaired active memory store $\mathcal{M}'$.

%% file: sections/04_benchmark.tex
\section{Controlled Memory-Repair Benchmark}
\label{sec:controlled-benchmark}

\paragraph{Task Construction}
We create 150 tool-use cases in shopping assistance, travel booking, and customer support, drawing on task structure from common agent benchmarks~\cite{yao2024tau,deng2023mind2web,xue2025illusion}. Each case contains a multi-turn session, memory store, tools, and paired clean and faulty traces. Following Section~\ref{sec:problem-definition}, every method receives $(\mathcal{U},\mathcal{M},\mathcal{T},\mathcal{F})$ and the available tools; the clean trace, expected repaired answer, benign memory labels, and node-level affected-state annotations are used only for evaluation. Prompt contracts used to instantiate agent executions are in Appendix~\ref{app:prompt}.

\paragraph{Fault Injection}

We inject four fault types motivated by observed memory risks~\cite{sunil2026memory,zou2025poisonedrag,hatalis2023memory,chao2026stale,chen2025halumem,zhang2026useful,chen2024agentpoison}: \emph{poisoned} records contain adversarial or corrupted values; \emph{stale} records survive a failed deletion, update or consolidation; \emph{wrong-user} records are associated with another user or context; and \emph{summary-drift} corrupts consolidation output. A case is retained only if injection changes the final answer and creates at least one downstream effect. Thus, the benchmark evaluates recovery after diagnosed faults; it does not estimate natural fault prevalence or detector accuracy.
The detailed fault injection strategies are provided in Appendix~\ref{app:fault-injection}.

Table~\ref{tab:controlled_stats} summarizes the benchmark distribution across domains and primary memory fault types. Each task is counted once under the fault type defining its benchmark stratum. Of the 150 tasks, 128 contain a single diagnosed faulty memory, while 22 contain two or three diagnosed faulty memories: 8 in customer support, 7 in shopping, and 7 in travel.

\begin{table}[t]
\centering
\small
\resizebox{\columnwidth}{!}{
\begin{tabular}{lrrrrr}
\hline
Domain & Pois. & Stale & W-user & Drift & Total \\
\hline
Shopping & 14 & 10 & 11 & 13 & 48 \\
Travel & 12 & 12 & 12 & 13 & 49 \\
Customer Support & 16 & 12 & 11 & 14 & 53 \\
\hline
Total & 42 & 34 & 34 & 40 & 150 \\
\hline
\end{tabular}
}
\caption{\textbf{Controlled benchmark statistics by domain and primary memory fault type.} Each task is counted once under the fault type defining its benchmark stratum. Pois. and W-user denote poisoned and wrong-user faults, respectively, while Drift denotes summary-drift.}
\label{tab:controlled_stats}
\end{table}

\paragraph{Evaluation Metrics}

We measure (i) \emph{recovery}, success under a deterministic task oracle that requires all expected answer facts and no fault-specific facts; (ii) \emph{recurrence}, the fraction of recovered cases whose answer fails the oracle again when the last user input is re-run on a fresh runtime loaded with the repaired store; (iii) removal of diagnosed faulty memories and preservation of benign memories; (iv) claim invalidation F1, micro-averaged between the gold affected claim steps of the faulty trace and the claim steps the repair actually invalidates; and (v) replayed-step ratio and LLM calls. Detailed metric definitions are provided in Appendix~\ref{app:metrics}.

%% file: sections/05_experiments.tex
\section{Experiments}

\begin{table*}[!t]
\centering
\small
\setlength{\tabcolsep}{3pt}
\resizebox{\textwidth}{!}{%
\begin{tabular}{lrrrrrrr}
\hline
Method & Recovery $\uparrow$ & Recurrence $\downarrow$ & Faulty removal $\uparrow$ & Benign preservation $\uparrow$ & Claim-inv.\ F1 $\uparrow$ & Replay ratio $\downarrow$ & LLM count $\downarrow$ \\
\hline
No repair & 0.000 & -- & 0.000 & 1.000 & 0.000 & 0.000 & 0.00 \\
Full memory reset & 0.407 & 0.197 & 1.000 & 0.000 & 0.000 & 0.061 & 4.03 \\
Delete retrieved memories & 0.340 & 0.431 & 0.881 & 0.789 & 0.000 & 0.061 & 4.05 \\
MemAudit-style & 0.320 & 0.292 & 0.311 & 0.980 & 0.000 & 0.051 & 10.43 \\
\hline
LLM-judge repair & 0.773 & 0.121 & 1.000 & 1.000 & 0.946 & 0.217 & 9.80 \\
AgentTrace-style & 0.607 & 0.341 & 0.000 & 0.997 & 0.853 & 0.091 & 4.69 \\
Ours & 0.853 & 0.266 & 1.000 & 1.000 & 0.566 & 0.123 & 5.70 \\
\hline
\end{tabular}%
}
\caption{\textbf{Main results on the controlled benchmark.} All metrics except LLM count are reported on a 0--1 scale. Recurrence is computed only over successfully recovered cases; ``--'' indicates that no case was recovered. Claim-inv.\ F1 denotes claim invalidation F1.}
\label{tab:main_results}
\end{table*}

\begin{table*}[!t]
\centering
\small
\setlength{\tabcolsep}{3pt}
\resizebox{\textwidth}{!}{%
\begin{tabular}{lrrrrrrr}
\hline
Variant & Recovery $\uparrow$ & Recurrence $\downarrow$ & Faulty removal $\uparrow$ & Benign preservation $\uparrow$ & Claim-inv.\ F1 $\uparrow$ & Replay ratio $\downarrow$ & LLM count $\downarrow$ \\
\hline
Full method & 0.853 & 0.266 & 1.000 & 1.000 & 0.566 & 0.123 & 5.70 \\
w/o support check & 0.880 & 0.265 & 1.000 & 0.986 & 0.507 & 0.154 & 6.51 \\
w/o rollback planner & 0.713 & 0.430 & 1.000 & 0.996 & 0.507 & 0.152 & 9.19 \\
w/o selective replay & 0.840 & 0.071 & 1.000 & 1.000 & 0.566 & 0.755 & 24.01 \\
\hline
\end{tabular}%
}
\caption{\textbf{Ablation results on the controlled benchmark.} The rollback planner mainly improves recovery and recurrence, support checking improves selectivity, and selective replay reduces repair cost.}
\label{tab:ablation}
\end{table*}

\begin{table*}[!t]
\centering
\small
\setlength{\tabcolsep}{3pt}
\resizebox{\textwidth}{!}{%
\begin{tabular}{lrrrrrrr}
\hline
Method & Recovery $\uparrow$ & Recurrence $\downarrow$ & Faulty removal $\uparrow$ & Benign preservation $\uparrow$ & Claim-inv.\ F1 $\uparrow$ & Replay ratio $\downarrow$ & LLM count $\downarrow$ \\
\hline
No repair & 0.000 & -- & 0.000 & 1.000 & 0.000 & 0.000 & 0.00 \\
Full memory reset & 0.060 & 0.000 & 1.000 & 0.000 & 0.000 & 0.062 & 4.84 \\
Delete retrieved memories & 0.100 & 0.400 & 0.949 & 0.466 & 0.000 & 0.062 & 4.26 \\
MemAudit-style & 0.260 & 0.231 & 0.577 & 0.969 & 0.000 & 0.062 & 16.52 \\
\hline
LLM-judge repair & 0.260 & 0.077 & 0.613 & 0.983 & 0.369 & 0.244 & 10.64 \\
AgentTrace-style & 0.540 & 0.037 & 0.000 & 0.997 & 0.603 & 0.123 & 6.98 \\
Ours & 0.680 & 0.059 & 1.000 & 0.993 & 0.669 & 0.179 & 8.22 \\
\hline
\end{tabular}%
}
\caption{\textbf{Transfer results on the adapted LongMemEval-V2 subset.} The subset contains 50 externally derived procedural tasks converted into our repair schema.}
\label{tab:lmev2_transfer}
\end{table*}

We evaluate all methods on the controlled benchmark of Section~\ref{sec:controlled-benchmark} and evaluate transfer on an adapted LongMemEval-V2 procedural subset. All methods use GPT-4o and share the same tool environment, task inputs, faulty memory store, failed execution trace, and diagnosed faulty memory identifiers; plan-based methods are executed by the same rollback executor. Additional Gemini-3.6-Flash and Qwen3.6-27B results are reported in Appendix~\ref{app:additional-result}. All results are from a single run at temperature 0, with seed 42 for GPT-4o and Qwen3.6-27B; Gemini-3.6-Flash does not expose a seed parameter. Qualitative analyses of representative repair cases in both settings are in Appendix~\ref{app:case-study}.

We compare against six baselines. Implementation details are in Appendix~\ref{app:baseline}. \textbf{No repair} leaves the faulty execution unchanged. Memory-centric baselines are \textbf{Full memory reset}, \textbf{Delete retrieved memories}, and a \textbf{MemAudit-style} baseline adapted from post-hoc memory auditing~\cite{tan2026memaudit}, which uses an oracle-assisted diagnostic fallback when its audit candidate set is empty and scores counterfactual influence with the task oracle; it therefore receives evaluation-side signals that other methods do not.
Trace-centric baselines are \textbf{LLM-judge repair}, inspired by Reflexion~\cite{shinn2023reflexion} and Self-Refine~\cite{madaan2023self}, and \textbf{AgentTrace-style}, adapted from causal graph tracing~\cite{wang2026agenttrace}. Prompts for LLM-judge repair are provided in Appendix~\ref{app:prompt}.

\subsection{Main Results}
\label{sec:main-results}

Table~\ref{tab:main_results} shows that our method achieves the highest recovery on the controlled benchmark. It recovers 85.3\% of cases, compared with 77.3\% for LLM-judge repair and 60.7\% for AgentTrace-style repair, absolute gains of 8.0 and 24.6 percentage points. Memory-centric baselines recover at most 40.7\%. Our method preserves 100.0\% of benign memories, versus 78.9\% for Delete retrieved memories and 0\% for full memory reset. Its 100\% faulty removal holds by construction, since the planner deletes every diagnosed fault ($\mathcal{M}_{\mathrm{del}}=\mathcal{F}$); LLM-judge repair is likewise required to delete them, whereas AgentTrace-style disables memory cleanup by design and therefore scores 0. Beyond these deletions, memory cleanup is rarely triggered: 141 of the 143 derived-memory candidates on the controlled benchmark retain trusted support, and no derived memory is quarantined (one on the adapted subset). The persistent-state repair measured here therefore consists mainly of deleting diagnosed faults and regenerating their producers.

Recurrence exposes the main limitation. Our method has lower recurrence than AgentTrace-style and Delete retrieved memories, but higher recurrence than LLM-judge repair (26.6\% vs.\ 12.1\%) and full memory reset (19.7\%). Thus removing diagnosed faults and identified unsupported state does not guarantee that the failure stays repaired when the store is reused; the gap is concentrated in stale faults (78.1\% recurrence, Appendix~\ref{app:additional-result}). The selective-replay ablation (Section~\ref{sec:ablation}) indicates that broader replay lowers recurrence at much higher cost. Because recurrence is computed only over recovered cases, it should be read together with recovery.

Memory-centric baselines directly edit memory state, but they either delete too much benign memory or fail to remove
downstream contamination. Full memory reset removes faulty memories but destroys all benign memory. Delete retrieved memories preserves more benign state but recovers far fewer cases than our method. And MemAudit-style repair is conservative but misses many faulty memories and derived effects. In contrast, trace-centric methods more accurately identify claim-level trace state requiring repair. LLM-judge repair and AgentTrace-style obtain higher claim invalidation F1 than our method on the controlled benchmark. However, this does not translate into better end-to-end repair. Compared with LLM-judge repair, our method achieves higher recovery while
reducing replay ratio by 43.3\% and LLM calls by 41.8\%. AgentTrace-style is cheaper, but this comes with a 24.6-percentage-point absolute drop in recovery and weaker faulty memory removal. Overall, our method offers a favorable recovery--cost trade-off: compared with LLM-judge repair, it achieves higher recovery with fewer replayed steps and LLM calls; compared with AgentTrace-style, it substantially improves recovery and faulty memory removal at the cost of additional replay and LLM calls.

\subsection{Ablation Study}
\label{sec:ablation}

The variants are defined as follows. \emph{W/o support check} marks every checkable affected step and derived-memory candidate as unsupported. \emph{W/o rollback planner} replays every affected claim, plan, final answer, memory mutation, and tool action, quarantines every derived-memory candidate, and performs neither support-based preservation nor answer-relevance selection. \emph{W/o selective replay} keeps the planner's memory dispositions but replays every replayable step of the trace, i.e., all memory reads, plans, tool actions, claims, memory mutations, and the last final answer; tool observations and earlier answers are not replay targets, so its replay ratio is 0.755 rather than 1.

In Table~\ref{tab:ablation}, removing the rollback planner produces the largest recovery drop among the tested ablations: it reduces recovery from 85.3\% to 71.3\%, increases recurrence from 26.6\% to 43.0\%, and lowers benign memory preservation from 100.0\% to 99.6\%. It also raises replay ratio from 12.3\% to 15.2\% and LLM calls from 5.70 to 9.19. Equivalently, the full method uses 38.0\% fewer LLM calls than the variant without the rollback planner. Within this ablation, rule-guided planning is associated with both higher recovery and fewer repair operations.

Support checking primarily improves preservation and selectivity rather than raw recovery. Removing it increases recovery from 85.3\% to 88.0\% and slightly reduces recurrence from 26.6\% to 26.5\%, but lowers benign memory preservation from 100.0\% to 98.6\%. It also increases replay ratio by 25.2\% and LLM calls by 14.2\%. Thus, support checking introduces a small recovery trade-off in exchange for preserving more independently supported state and reducing unnecessary recomputation.

Selective replay primarily controls repair cost. Without it, recovery decreases slightly from 85.3\% to 84.0\%, while recurrence improves from 26.6\% to 7.1\%. This reduction in recurrence, however, requires substantially broader replay: replay ratio rises from 12.3\% to 75.5\%, and LLM calls increase from 5.70 to 24.01. Relative to this broader-replay variant, selective replay reduces replay ratio by 83.7\% and LLM calls by 76.3\%. Selective replay is therefore an efficiency mechanism with an explicit recurrence--cost trade-off, enabling targeted repair rather than exhaustive recomputation.

\subsection{Transfer to Adapted LongMemEval-V2 Subset}
\label{sec:lmev2-transfer}

We evaluate an adapted subset of LongMemEval-V2~\cite{wu2026longmemeval}. The subset contains 50 externally derived procedural and navigation-style tasks converted into our repair schema (Appendix~\ref{app:adapted-lmev2}).
Unlike the controlled benchmark, this subset targets the poisoned faulty type and is predominantly multi-fault: 5 cases contain one faulty memory, while 45 contain 2--4 faults, as summarized in Table~\ref{tab:lmev2-fault-distribution}.
We therefore treat it as an end-to-end transfer stress test on externally derived multi-fault trajectories rather than a fault-type coverage study.

Table~\ref{tab:lmev2_transfer} shows that our method again achieves the highest recovery, 68.0\% against 54.0\% for AgentTrace-style and 26.0\% for LLM-judge repair. In contrast to its controlled benchmark result, our method
also obtains the highest claim invalidation F1 on this subset, scoring 0.669 versus 0.603 and 0.369. One possible explanation is that these trajectories are dominated by poisoned memories on the answer path, so the claim steps requiring repair may largely overlap with the answer-relevant region selected for replay. It also ties for the highest faulty memory removal and preserves 99.3\% of benign memories. AgentTrace-style keeps a cost advantage and has slightly better recurrence (3.7\% vs.\ 5.9\%) and preservation (99.7\% vs.\ 99.3\%), but trails by 14.0 points in recovery; against LLM-judge repair, our method is better on every reported transfer metric while using 73.4\% of its replay ratio and 77.3\% of its LLM calls. The lower recovery on the adapted subset (68.0\% vs.\ 85.3\%) indicates sensitivity to the evaluation setting, which differs in domain, fault type, and fault count; neither result should be interpreted as a deployment estimate.

%% file: sections/06_conclusion.tex
\section{Conclusion}

This paper studies post-failure memory recovery for memory-augmented agents. We propose dependency-guided rollback repair, which traces contamination through a typed memory-to-action graph, preserves independently supported state, constructs rule-guided rollback plans, and selectively replays answer-relevant affected steps. We also build a controlled benchmark spanning shopping, travel, and customer support. Across both evaluations, our method achieves the highest immediate recovery and a favorable recovery--cost trade-off.

\paragraph{Limitations.}
Our results hold under several assumptions. Faulty memories are diagnosed upstream, and the agent runtime must emit provenance for memory reads, claims, plans, actions, observations, and mutations; missing edges can under-trace contamination, while spurious edges can cause unnecessary invalidation. Rollback repairs agent-maintained state only, and replaying side-effecting tools requires resettable or compensating interfaces. The controlled benchmark is constructed by fault injection and does not estimate natural fault prevalence, and the adapted LongMemEval-V2 subset covers only poisoned, mostly multi-fault procedural cases. All results come from a single run per configuration. Finally, selective replay leaves higher recurrence than broader recomputation on the controlled benchmark, especially for stale faults; improving recurrence robustness and claim-state identification without sacrificing selective repair is the main direction for future work.

%% file: appendix.tex
\appendix

\setcounter{secnumdepth}{2}
\setcounter{section}{0}
\renewcommand{\thesection}{\Alph{section}}

\input{appendix/data_schema}

\input{appendix/prompt}

\input{appendix/rollback_rules}       %

\input{appendix/fault_injection}

\input{appendix/metrics}

\input{appendix/additional_result}

\input{appendix/case_study}

\input{appendix/baseline}

\input{appendix/adapted_lmev2}

%% file: appendix/data_schema.tex
\section{Dependency-guided Rollback Repair Data Schema}
\label{app:data-schema}

We provide the data schemas used by our implementation. Each data instance contains a user session record, a memory store, an execution trace, diagnosed faulty memories, a corresponding dependency graph, and a repair plan generated during method execution. These artifacts correspond to the modules in the main paper: the graph construction module uses lineage fields, cited identifiers, generated memory identifiers, and trace predecessor/successor fields to build typed dependency edges; the rollback planner traverses the graph from diagnosed faulty memories to identify affected claims, downstream memories, and replay candidates; and the plan executor consumes the rollback plan to produce the repaired memory store and repaired trace.

The clean store is used only as an evaluation reference, the faulty store is the input to the repair method, and the repaired store is the output produced by the plan executor. These stores share the same memory record schema. Identifiers are preserved for unchanged memories.

\paragraph{Memory record.}
As shown in Table~\ref{tab:schema_memory}, each memory record contains a memory identifier, textual content, status, source metadata, lineage metadata, and an optional trust score.
In our current implementation, the trust score is recorded for analysis only, not used by the repair algorithm. Future work could estimate it more robustly and integrate it into the support checking module.

\begin{table}[b!]
\centering
\small
\begin{tabularx}{\columnwidth}{l X}
\toprule
Field & Description \\
\midrule
\texttt{memory\_id} & Unique memory identifier. \\
\texttt{task\_id} & Task to which the memory belongs. \\
\texttt{content} & Natural language memory content. \\
\texttt{source} & Source of the memory. \\
\texttt{last\_modified\_by} & User input or step that last modified the memory. \\
\texttt{fault\_type} & Fault label if the memory is faulty. \\
\texttt{created\_at} & Creation timestamp. \\
\texttt{last\_modified\_at} & Last modification timestamp. \\
\texttt{derived\_from} & Ancestor memories used to derive this memory. \\
\texttt{supersedes} & Prior memory superseded by this memory. \\
\texttt{trust\_score} & Trust score assigned to the memory. \\
\texttt{status} & Memory state, one of active, deleted, quarantined, or superseded. \\
\texttt{fact\_key} & Structured fact key represented by the memory. \\
\texttt{fact\_value} & Structured fact value represented by the memory. \\
\texttt{entity\_id} & Entity associated with the fact. \\
\bottomrule
\end{tabularx}
\caption{Schema of memory records in clean, faulty, and repaired memory stores.
In our implementation, \texttt{source} takes values from user inputs, tool observations, agent summaries, claims, and injected faults, while
\texttt{fault\_type} covers poisoned, stale, wrong-user, and summary-drift memories.}
\label{tab:schema_memory}
\end{table}

\paragraph{Execution step.}
Table~\ref{tab:schema_trace_step} summarizes the fields in each execution step, including the step identifier, semantic type, textual content, cited memories or prior steps, tool metadata when applicable, and memory mutation metadata when applicable.

\begin{table}[t!]
\centering
\small
\begin{tabularx}{\columnwidth}{l X}
\toprule
Field & Description \\
\midrule
\texttt{step\_id} & Unique execution step identifier. \\
\texttt{task\_id} & Task to which the step belongs. \\
\texttt{step\_type} & Semantic step type. \\
\texttt{content} & Natural language content of the step. \\
\texttt{prev\_step\_ids} & Predecessor steps in the trace. \\
\texttt{next\_step\_ids} & Successor steps in the trace. \\
\texttt{invalidated\_memory\_ids} & Memories invalidated by this step. \\
\texttt{generated\_memory\_ids} & Memories produced by this step. \\
\texttt{used\_ids} & Memory or step identifiers used by this step. \\
\texttt{tool\_name} & Tool name for tool action steps. \\
\texttt{tool\_args} & Tool arguments for tool action steps. \\
\texttt{timestamp} & Step timestamp. \\
\bottomrule
\end{tabularx}
\caption{Schema of execution steps in clean, faulty, and repaired traces.
\texttt{step\_type} covers memory reads, claims, planning steps, tool actions and observations, final answers, and memory write/delete/update/consolidation
operations.}
\label{tab:schema_trace_step}
\end{table}

\paragraph{Dependency graph.}
Each graph node has a node identifier, node kind, semantic type, and metadata, and each graph edge has a source node, target node, and one or more typed labels. Detailed fields are presented in Tables~\ref{tab:schema_graph_node} and \ref{tab:schema_graph_edge}, with edge labels listed in Table~\ref{tab:schema_graph_edge_label}.

The graph is constructed from the memory store and execution trace, so memory nodes correspond to memory records, step nodes correspond to execution step records, and edge endpoints resolve to nodes in the same task instance. Chain edges are consistent with \texttt{prev\_step\_ids} and \texttt{next\_step\_ids}; citation and support edges are consistent with \texttt{used\_ids}, while production edges are consistent with \texttt{generated\_memory\_ids}. Edge directions follow provenance and execution flow. Supporting or antecedent nodes point to dependent nodes, and earlier execution steps point to later steps.

\begin{table}[t!]
\centering
\small
\begin{tabularx}{\columnwidth}{l X}
\toprule
Field & Description \\
\midrule
\texttt{node\_id} & Unique graph node identifier. \\
\texttt{node\_kind} & Node kind, one of memory, step, or user\_input. \\
\texttt{semantic\_type} & Semantic category used by the repair pipeline, including memory, user\_input, or one of the execution step types in Table~\ref{tab:schema_trace_step}.\\
\texttt{step\_type} & Original trace step type for step nodes; same vocabulary as execution step \texttt{step\_type}, and null for non-step nodes. \\
\texttt{memory\_status} & Memory status for memory nodes; same vocabulary as \texttt{status}, and null for non-memory nodes. \\
\texttt{raw} & Original raw record used to construct the node. \\
\bottomrule
\end{tabularx}
\caption{Schema of dependency graph nodes.}
\label{tab:schema_graph_node}
\end{table}

\begin{table}[t!]
\centering
\small
\begin{tabularx}{\columnwidth}{l X}
\toprule
Field & Description \\
\midrule
\texttt{source\_id} & Source node identifier in the adjacency map. \\
\texttt{target\_id} & Target node identifier in the adjacency map. \\
\texttt{labels} & Edge labels defined in Table~\ref{tab:schema_graph_edge_label}. \\
\bottomrule
\end{tabularx}
\caption{Logical schema of dependency graph edges. In the implementation, edges are stored as adjacency maps from source nodes to target nodes with a list of typed labels.}
\label{tab:schema_graph_edge}
\end{table}

\begin{table}[t]
\centering
\small
\begin{tabularx}{\columnwidth}{l X}
\toprule
Label & Meaning \\
\midrule
\texttt{chain} & Step $\rightarrow$ step; execution or reasoning order. \\
\texttt{cite} & Cited node $\rightarrow$ citing step; explicit citation relation. \\
\texttt{support} & Memory $\rightarrow$ step; the memory supports a step. \\
\texttt{produce} & Mutating step $\rightarrow$ memory; the step writes a memory. \\
\texttt{delete} & Memory $\rightarrow$ memory delete step. \\
\texttt{update} & Memory $\rightarrow$ memory update step. \\
\texttt{consolidate} & Memory $\rightarrow$ memory consolidate step. \\
\texttt{supersede} & Old memory $\rightarrow$ superseding memory. \\
\texttt{derive} & Ancestor memory $\rightarrow$ derived memory. \\
\texttt{initiate} & User input $\rightarrow$ first step of the turn. \\
\bottomrule
\end{tabularx}
\caption{Dependency graph edge labels, including their direction and meaning.}
\label{tab:schema_graph_edge_label}
\end{table}

\paragraph{Repair plan.}
Repair plans contain task identifiers and detailed instructions for the plan executor in Table~\ref{tab:schema_rollback_plan}. Memories are deleted or quarantined, while selected steps are replayed. Invalidated claims and redundant steps support state reconstruction, and suspicious steps are retained for audit rather than replayed by the default executor.

Plan fields are interpreted as disjoint action sets unless otherwise specified: a memory cannot appear in both \texttt{delete\_memory\_ids} and \texttt{quarantine\_memory\_ids}, and a trace step cannot simultaneously appear in \texttt{replay\_step\_ids}, \texttt{preserve\_step\_ids}, and \texttt{redundant\_step\_ids}. All identifiers in the plan must resolve to memories or trace steps in the same task instance.

\begin{table}[t]
\centering
\small
\begin{tabularx}{\columnwidth}{l X}
\toprule
Field & Description \\
\midrule
\texttt{task\_id} & Task identifier. \\
\texttt{delete\_memory\_ids} & Memories selected for deletion. \\
\texttt{quarantine\_memory\_ids} & Memories selected for quarantine. \\
\texttt{invalidate\_claim\_ids} & Claims or trace steps selected for invalidation. \\
\texttt{replay\_step\_ids} & Trace steps selected for replay. \\
\texttt{preserve\_step\_ids} & Trace steps explicitly preserved. \\
\texttt{redundant\_step\_ids} & Steps considered redundant after planning. \\
\texttt{suspicious\_step\_ids} & Suspicious but non-replayed steps. \\
\bottomrule
\end{tabularx}
\caption{Schema of rollback plan artifacts.}
\label{tab:schema_rollback_plan}
\end{table}

%% file: appendix/prompt.tex
\section{Prompt Contracts and Templates}
\label{app:prompt}

\begin{table*}[t!]
\centering
\small
\begin{tabularx}{\textwidth}{l X}
\toprule
Domain/source & Domain prompt \\
\midrule
Shopping & You are a helpful shopping assistant. You can help users find products, compare prices, and provide recommendations based on their preferences. \\
Travel & You are a helpful travel assistant. You can help users search for flights and hotels, check availability and prices, and provide recommendations based on their travel preferences. \\
Customer support & You are a helpful customer support assistant. You can help users look up orders, check account or ticket status, troubleshoot issues, and provide recommendations based on their support history. \\
\hline
Adapted LongMemEval-V2 subset & You are a helpful long-memory assistant. Use the provided memories and prior interaction history to answer procedural or navigation-style user requests. Follow the user's stored preferences and constraints, and use available tools only when they are needed to complete the requested procedure. \\
\bottomrule
\end{tabularx}
\caption{Domain instructions used to instantiate agent prompts.}
\label{tab:domain-prompts}
\end{table*}

\begin{table*}[t!]
\centering
\small
\setlength{\tabcolsep}{4pt}
\begin{tabularx}{\textwidth}{p{0.17\textwidth} X p{0.22\textwidth}}
\toprule
Prompt & Key instruction summary & Output fields \\
\midrule
Claim generation & Produce a concise claim for the user's current intent from the request, retrieved memories, and prior turns; cite all supporting memory, prior turn, and current input ids. & Claim; cited evidence \\
Tool planning & Plan only valid tool calls. Use read-only tools for external or volatile information and side-effecting tools only under an explicit current instruction. Ground tool arguments in available evidence, avoid invented ids, repeated failed calls, and duplicates. & Tool steps; tool name; arguments; cited evidence \\
Response generation & Answer from the claim, memories, and tool results. The response must be grounded, concise, and faithful to observed tool outcomes: it must not invent unavailable data, ignore returned results, or report a side-effecting action as completed unless this turn's tool result supports it. & Response; cited evidence \\
Memory mutation & Propose durable memory writes, updates, deletes, or consolidations. Memories encode atomic facts; tool observation memories are immutable; every operation records provenance so later repair can identify the memories and trace steps supporting each mutation. & New memories; updates; deletes; consolidations; provenance \\
\bottomrule
\end{tabularx}
\caption{Prompt contracts for the base memory-augmented agent.}
\label{tab:agent-prompts}
\end{table*}

\begin{table*}[t!]
\centering
\small
\setlength{\tabcolsep}{4pt}
\begin{tabularx}{\textwidth}{p{0.19\textwidth} X p{0.22\textwidth}}
\toprule
Prompt & Key instruction summary & Output fields \\
\midrule
Missing action replan & Reuse the planning contract over the repaired active state to decide whether additional tool calls are still required; empty plans are valid and tool names must come from the available tool set. & Tool steps; cited evidence \\
Claim and answer replay & Regenerate the claim or final answer from repaired memories, repaired history, and repaired tool observations, citing only evidence visible in the repaired context. & Claim or response; cited evidence \\
Targeted tool action replay & Re-derive one replacement call for each superseded tool action. Old calls and observations are untrusted hints; replacements may change tool name or arguments when needed, but may not introduce extra unbound calls. & Target action id; tool name; arguments; cited evidence \\
Targeted memory replay & Re-derive one replacement for each superseded memory mutation step using repaired evidence while preserving the target semantic type and avoiding protected preserved memory outputs. & Target step id; content; source; trust score; cited evidence \\
\hline
LLM-judge repair baseline & Predict a rollback plan from compact faulty artifacts, including faulty memories, faulty trace steps, session turns, the wrong final answer, valid ids, and the executor plan contract. The prompt requires strict JSON, forbids invented ids, deletes source faulty memories, and requests a minimal dependency complete replay region. & Delete; quarantine; invalidate; replay; preserve; redundant; suspicious; reason \\
\bottomrule
\end{tabularx}
\caption{Prompt contracts for rollback replay and the LLM-judge repair baseline.}
\label{tab:repair-prompts}
\end{table*}

We summarize the structured prompt contracts used by the base agent, our rollback executor, and the LLM-judge repair baseline. Because these components are implemented through structured prompting, the contracts are part of the experimental protocol. They define what information each component can access, what evidence it must cite, and what JSON outputs it must produce. Runtime prompts instantiate these reusable contracts with domain instructions, user turns, retrieved memories, prior turns, tool schemas, tool observations, repaired state artifacts, and required output schemas. We report the contracts rather than fully instantiated prompts.

\paragraph{Domain Prompts.}
The executor prepends one domain instruction to the claim, planning, response, and memory mutation prompts. Table~\ref{tab:domain-prompts} lists the controlled benchmark instructions and the separate instruction used for the adapted LongMemEval-V2 subset. These domain prompts are intentionally concise so that domain framing does not encode repair specific behavior.

\paragraph{Agent Prompts.}
The base memory-augmented agent decomposes each turn into claim generation, planning and tool use, response generation, and post-execution memory mutation. Each stage returns strict JSON with provenance identifiers, which are recorded in the trace and later used by graph construction and repair. The planning output is parsed by the executor into concrete tool action steps. Corresponding tool results are recorded as tool observation steps and made available to response generation and memory mutation.

\paragraph{Rollback Repair and Baseline Prompts.}
Replay prompts are selected by the rollback plan and instantiated over the repaired active state. They return strict JSON with evidence ids. For targeted replay, old actions or memory mutations are shown only as untrusted hints, and the model must produce replacements grounded in repaired evidence.

%% file: appendix/rollback_rules.tex
\section{Rule-Guided Rollback Decisions}
\label{app:rollback-rules}

\begin{table*}[!b]
\centering
\small
\setlength{\tabcolsep}{6pt}
\renewcommand{\arraystretch}{1.12}
\begin{tabularx}{\textwidth}{
    @{}
    >{\raggedright\arraybackslash}p{0.35\textwidth}
    >{\raggedright\arraybackslash}X
    @{}
}
\toprule
\textbf{Condition} & \textbf{Rollback rule} \\
\hline
Diagnosed faulty memory
& Delete it from the active memory store. \\

Unsupported derived memory carrying a fault-specific fact
& Quarantine it and exclude it from subsequent retrieval. \\
\hline
Candidate supported by independent evidence
& Preserve the existing trace output, together with any unchanged downstream action or observation that depends on it. \\

Unsupported answer-relevant execution node
& Invalidate the old output and replay the corresponding executable step. \\

Unsupported non-answer-relevant execution node
& Invalidate the old output without replay and exclude it from subsequent context. \\

Diagnosed faulty-memory provenance seed
& If a provenance seed exists, replay the step that produced the faulty memory to generate a corrected replacement. \\

Final-answer node
& Always regenerate the final answer under the repaired memory and trace state. \\
\bottomrule
\end{tabularx}
\caption{\textbf{Rule-guided rollback decisions.} Memory dispositions determine which records remain active, whereas trace decisions determine which prior outputs are preserved, invalidated, or replayed.}
\label{tab:rollback-rules}
\end{table*}

Table~\ref{tab:rollback-rules} summarizes the principal repair rules applied by the rollback planner. %
Memory dispositions determine which records remain active, while trace actions determine which prior outputs are invalidated, replayed, or preserved.

%% file: appendix/fault_injection.tex
\section{Fault Injection Details}
\label{app:fault-injection}

We inject faulty memories into clean benchmark instances to construct post-failure memory states. Fault selection is manifest-driven: for each task, a manifest specifies one or more fault specifications, including the fault type, target memory or target mutation step when applicable, and optional corrupted content. The injector does not score candidate memories or run the agent. Instead, it rewrites the clean memory store and cut trace so that the replay starts from a seeded faulty state. When multiple faults are specified for the same task, they are applied sequentially and merged into a single faulty instance.

Each manifest row contains a task identifier and an ordered list of fault specifications:
\[
r=(\tau, [f_1,\ldots,f_k]).
\]
Here, $\tau$ denotes the task identifier, and faults are applied sequentially in the listed order. Each fault specification $f_i$ records the fault type and any fault specific fields needed for injection, such as the target memory, target mutation step, corrupted content, or wrong-user content. Corrupted content is used for manifest specified corruptions, while wrong-user content is used only for wrong-user faults.

\paragraph{Poisoned.}
For poisoned memory faults, the manifest names a target memory produced by a memory write step. The injector replaces the clean memory content with manifest provided corrupted content when available; otherwise, it corrupts a concrete value using deterministic domain vocabularies or numeric shifts. The faulty record receives a new faulty identifier, is marked active, and keeps the original temporal position and source metadata while being stamped with injected fault provenance. The producing trace step is also rewritten so that its generated memory id and textual content are consistent with the corrupted memory.

For example, in a shopping task, a clean memory such as ``The user prefers black running shoes'' may be corrupted into ``The user prefers red running shoes.'' The corresponding memory write step is rewritten to generate the faulty memory id and the corrupted content, so the seeded trace and memory store remain consistent.

\paragraph{Stale.}
To model stale memory faults, the injector simulates a failed memory mutation in which an old memory that should have been deleted or superseded remains active. It identifies the delete, update, or consolidation step that consumed the target memory specified in the manifest. The consumed memory inputs are reactivated as stale faulty records, while the new memories that would have been produced by the mutation, together with their downstream lineage when applicable, are removed from the seeded store. The consuming step is kept in the cut trace but rewritten so that generated memories are removed and references point to the surviving stale records.

A clean trace may update a memory stating ``The user prefers economy flights'' into a new memory stating ``The user prefers business class flights.'' A stale injection reactivates the old economy flight preference and removes the new business class preference from the seeded store. During replay, this causes the agent to behave as if the user still prefers economy flights.

\paragraph{Wrong-user.}
Wrong-user faults are injected by fabricating a new active memory from the manifest provided wrong-user content. This memory has no producing trace step and is treated as an injected root memory store fault. It is inserted at the beginning of the memory timeline, and replay starts from the first turn so that the wrong-user memory can influence subsequent execution.

For example, a travel task for the current user may receive an injected memory stating ``The user prefers hotels with airport shuttle service,'' even though this preference belongs to a different user. Since the injected memory has no producing step in the current trace, it is treated as a root memory store fault and can affect replay from the first turn.

\paragraph{Summary-drift.}
Summary-drift faults use the same injection mechanism as poisoned memories, but they specifically target derived memories produced by summarization or consolidation. This models an inaccurate summary or consolidation output whose content differs from the clean memory while appearing as a confident derived memory. When specified, additional derived memories can be dropped so that the seeded store reflects the drifted summary lineage.

In a customer support scenario, two clean memories may state that ``the user reported defective headphones'' and ``the user asked about refund eligibility after returning the item.'' A correct consolidation should summarize them as ``The user returned defective headphones and is seeking a refund,'' but summary-drift changes the derived memory to ``The user wants to keep the headphones and receive a replacement.'' This drifted summary may cause the agent to create the wrong support ticket or choose an incorrect resolution policy.

%% file: appendix/metrics.tex
\section{Metrics Details}
\label{app:metrics}

We evaluate each repair method at the case level and report ratio-based metrics on a 0--1 scale unless otherwise stated. Memory, claim, and replay metrics use micro-aggregation: their numerators and denominators are summed across cases before the final ratio is computed, thereby avoiding equal weighting of cases containing different numbers of memories, claims, or trace steps.

\paragraph{Recovery.} Recovery measures whether the repaired agent produces a correct final outcome according to a deterministic task oracle. For the controlled shopping, travel, and customer support tasks, a case succeeds if the final answer contains all required facts and none of the fault-specific facts. For adapted tasks, each case carries an evaluator specification derived from LongMemEval-V2, and we apply a deterministic port of its phrase-set or multiple-choice answer matching; the case must additionally include any expected tool calls and use the memories its answer depends on. None of the 50 retained cases uses an LLM-judged evaluator: 34 use phrase-set matching, 10 ordered phrase-set matching, and 6 multiple-choice matching. Let $N_{\mathrm{succ}}$ and $N_{\mathrm{fail}}$ be the numbers of cases labeled as success and failure, respectively. We compute
\[
\mathrm{Recovery}
=
\frac{N_{\mathrm{succ}}}
{N_{\mathrm{succ}}+N_{\mathrm{fail}}}.
\]

\paragraph{Recurrence.} Recurrence measures whether a repaired state remains correct when it is reused and is evaluated only on cases that pass the recovery oracle. For each successfully recovered case, we instantiate a fresh agent runtime, clear its existing state, load only the active memories from the repaired memory store, and rerun the final user input. The outcome is labeled as recurrence if a fault-specific fact reappears in the answer or an expected fact is missing; probes that cannot be judged deterministically are labeled uncertain and excluded. Let $N_{\mathrm{rec}}$ and $N_{\mathrm{no\text{-}rec}}$ denote the numbers of decided recurrence and no-recurrence probes, respectively. We report
\[
\mathrm{Recurrence}
=
\frac{N_{\mathrm{rec}}}
{N_{\mathrm{rec}}+N_{\mathrm{no\text{-}rec}}}.
\]
Because recurrence is conditioned on successful recovery, the two metrics should be interpreted jointly.

\paragraph{Faulty removal.} Faulty removal measures the recall of source faulty memories removed by the repair method. For case $c$, let $F_c$ be the set of source faulty memory ids specified by the ground truth, and let $D_c$ contain memories deactivated by the repair or selected for deletion or quarantine:
\[
D_c
=
D_c^{\mathrm{deactivated}}
\cup D_c^{\mathrm{delete}}
\cup D_c^{\mathrm{quarantine}}.
\]
We compute the micro-aggregated score
\[
\mathrm{FaultyRemoval}
=
\frac{\sum_c |F_c\cap D_c|}
{\sum_c |F_c|}.
\]
Thus, the metric credits repair-attributable removal decisions rather than merely checking whether a memory is absent from the final active store. AgentTrace-style achieves a Faulty Removal score of 0 because it focuses exclusively on trace repair and leaves the memory store unchanged.

\paragraph{Benign preservation.} Benign preservation measures how much non-faulty memory state survives the repair. For case $c$, let $B_c$ denote the ground-truth benign memory ids. A benign memory that was active when repair began is considered preserved if either its original id or a successor reachable through the memory id remapping chain remains active after repair. A benign memory that was already inactive is considered preserved as long as the repair does not target it for deactivation, deletion, or quarantine. If $P_c\subseteq B_c$ is the set of preserved benign memories, we compute
\[
\mathrm{BenignPreservation}
=
\frac{\sum_c |P_c|}
{\sum_c |B_c|}.
\]

\paragraph{Claim-invalidation F1 ($\uparrow$).} Claim-invalidation F1 evaluates whether the repair invalidates exactly the trace claims affected by the fault. For case $c$, let $G_c$ be the ground-truth set of affected trace steps whose type is \texttt{claim}, and let $I_c$ be the set of claim ids actually invalidated by the repair. We first micro-aggregate
\[
\begin{aligned}
\mathrm{TP}&=\textstyle\sum_c |G_c\cap I_c|,\qquad
\mathrm{FP}=\textstyle\sum_c |I_c\setminus G_c|,\\
\mathrm{FN}&=\textstyle\sum_c |G_c\setminus I_c|,
\end{aligned}
\]
and then report
\[
\mathrm{ClaimInvF1}
=
\frac{2\mathrm{TP}}
{2\mathrm{TP}+\mathrm{FP}+\mathrm{FN}}.
\]
The score for memory-centric baselines is always 0 because they only focus on memory store repair and no claim steps are invalidated at all.

\paragraph{Replay ratio.} Replay ratio measures the fraction of the original trace that is actually regenerated. Let $S_c$ be the total number of steps in the original trace and $R_c$ the number of original target steps that are both selected for replay and successfully mapped to generated replacements. We report
\[
\mathrm{ReplayRatio}
=
\frac{\sum_c R_c}
{\sum_c S_c}.
\]
Auxiliary replacement steps generated as a consequence of replay, such as tool observations, do not increase the numerator unless they are themselves original replay targets. The metric therefore measures the selectively replayed portion of the original execution rather than the total length of the repaired trace. Although memory-centric methods modify only the memory store, they replay the final user turn to regenerate the response under the repaired memory state, resulting in a nonzero replay ratio. MemAudit-style has a lower ratio because it emits a no-op and skips final turn replay when no candidate memory is found.

\paragraph{LLM count.} LLM count reports the average number of LLM calls made by the repair procedure per case. For our method, the per-case count is
\[
L_c
=
L_c^{\mathrm{decision}}
+
L_c^{\mathrm{replay}},
\]
where the two terms count calls used to make repair decisions and calls used to regenerate replayed steps, respectively. For $N$ evaluated cases, we report
\[
\mathrm{LLMCount}
=
\frac{1}{N}\sum_{c=1}^{N}L_c.
\]
This count excludes calls made during the original faulty execution, deterministic evaluation, and subsequent recurrence probing.

%% file: appendix/additional_result.tex
\section{Additional Results}
\label{app:additional-result}

\begin{table*}[t!]
\centering
\small
\setlength{\tabcolsep}{4pt}
\begin{tabular*}{\textwidth}{@{\extracolsep{\fill}}lrrrrrr}
\hline
Domain & $n$ & Recovery $\uparrow$ & Recurrence $\downarrow$ & Claim-inv.\ F1 $\uparrow$ & Replay ratio $\downarrow$ & LLM count $\downarrow$ \\
\hline
Customer support & 53 & 0.943 & 0.240 & 0.442 & 0.155 & 7.04 \\
Shopping & 48 & 0.729 & 0.400 & 0.727 & 0.076 & 4.31 \\
Travel & 49 & 0.878 & 0.186 & 0.621 & 0.143 & 5.61 \\
\hline
\end{tabular*}
\caption{\textbf{Per-domain results for our method.} The method remains effective across all three domains. Shopping has the lowest recovery.}
\label{tab:app_domain_results}
\end{table*}

\begin{table*}[t!]
\centering
\small
\setlength{\tabcolsep}{4pt}
\begin{tabular*}{\textwidth}{@{\extracolsep{\fill}}lrrrrrr}
\hline
Fault type & $n$ & Recovery $\uparrow$ & Recurrence $\downarrow$ & Claim-inv.\ F1 $\uparrow$ & Replay ratio $\downarrow$ & LLM count $\downarrow$ \\
\hline
Poisoned & 42 & 0.810 & 0.000 & 0.341 & 0.116 & 5.83 \\
Stale & 34 & 0.941 & 0.781 & 0.635 & 0.154 & 5.53 \\
Wrong-user & 34 & 0.706 & 0.125 & 0.421 & 0.152 & 7.12 \\
Summary-drift & 40 & 0.950 & 0.158 & 0.870 & 0.084 & 4.50 \\
\hline
\end{tabular*}
\caption{\textbf{Per-fault-type results for our method.} Fault types are grouped by primary fault type. Summary-drift and stale faults have higher recovery, while wrong-user faults are harder to recover and require more LLM calls.}
\label{tab:app_fault_type_results}
\end{table*}

\begin{table*}[t!]
\centering
\small
\setlength{\tabcolsep}{3pt}
\resizebox{\textwidth}{!}{%
\begin{tabular}{llrrrrrrr}
\hline
Backbone & Method & Recovery $\uparrow$ & Recurrence $\downarrow$ & Faulty removal $\uparrow$ & Benign preservation $\uparrow$ & Claim-inv.\ F1 $\uparrow$ & Replay ratio $\downarrow$ & LLM count $\downarrow$ \\
\hline
GPT-4o & LLM-judge repair & 0.773 & 0.121 & 1.000 & 1.000 & 0.946 & 0.217 & 9.80 \\
GPT-4o & AgentTrace-style & 0.607 & 0.341 & 0.000 & 0.997 & 0.853 & 0.091 & 4.69 \\
GPT-4o & Ours & 0.853 & 0.266 & 1.000 & 1.000 & 0.566 & 0.123 & 5.70 \\
\hline
Gemini-3.6-Flash & LLM-judge repair & 0.740 & 0.108 & 1.000 & 0.987 & 0.604 & 0.183 & 8.37 \\
Gemini-3.6-Flash & AgentTrace-style & 0.513 & 0.156 & 0.000 & 0.997 & 0.853 & 0.091 & 4.64 \\
Gemini-3.6-Flash & Ours & 0.733 & 0.091 & 1.000 & 1.000 & 0.566 & 0.123 & 5.53 \\
\hline
Qwen3.6-27B & LLM-judge repair & 0.613 & 0.196 & 1.000 & 0.993 & 0.739 & 0.133 & 6.37 \\
Qwen3.6-27B & AgentTrace-style & 0.487 & 0.260 & 0.000 & 0.997 & 0.853 & 0.091 & 4.64 \\
Qwen3.6-27B & Ours & 0.660 & 0.061 & 1.000 & 1.000 & 0.566 & 0.123 & 5.52 \\
\hline
\end{tabular}%
}
\caption{\textbf{Backbone sensitivity on the controlled benchmark.} Supplemental Gemini-3.6-Flash and Qwen3.6-27B runs evaluate our method together with the closest LLM-based and trace-based repair competitors.}
\label{tab:app_backbone_results}
\end{table*}

This appendix reports additional controlled benchmark results. We focus on slices that explain where recovery remains difficult, how repair behavior changes across fault types, how results vary across LLM backbones, and how effectiveness relates to replay and LLM cost.

\paragraph{Per-domain results.}
Table~\ref{tab:app_domain_results} shows that performance varies across domains. Customer support achieves the highest recovery at 94.3\%, while shopping is the hardest domain, with recovery dropping to 72.9\%. Shopping also has the highest recurrence at 40.0\%, suggesting that repaired answers are more likely to remain vulnerable to downstream state errors in this domain. We attribute this difficulty to fine-grained product distinctions, including product ids, sellers, prices, inventory, and user preferences, where a small reconstruction error can change the selected item or seller.

\paragraph{Per-fault-type results.}
Table~\ref{tab:app_fault_type_results} groups tasks by primary fault type. Summary-drift has the highest recovery at 95.0\% and the highest claim invalidation F1 at 0.870, likely because drifted summaries often have compact provenance through update or consolidation steps. Wrong-user faults are the hardest, with recovery dropping to 70.6\% and the highest LLM count, reflecting that wrong-user memories can look like coherent personalized records unless the trace exposes a conflict with the current user's history. Stale faults show a different pattern. Recovery is high at 94.1\%, but recurrence is also high at 78.1\%, indicating that final answer repair can succeed while stale state suppression remains fragile. Poisoned faults have the lowest claim invalidation F1, but recurrence is 0 among recovered cases, suggesting that once the corrupted value is removed or refreshed, the repaired state is less likely to reintroduce the same failure.

\paragraph{Backbone sensitivity.}
Table~\ref{tab:app_backbone_results} reports supplemental controlled benchmark runs with Gemini-3.6-Flash and Qwen3.6-27B.
We run all three LLMs with temperature 0 and seed 42, except that Gemini-3.6-Flash does not accept a seed parameter.
The goal is to test whether the main findings depend on a particular LLM backbone, rather than to run the full comparison across all repair paradigms.
We therefore compare our method with LLM-judge repair, the closest LLM-based competitor, and AgentTrace-style, the closest trace-based competitor.
Compared with GPT-4o, recovery decreases for all three methods under the additional backbones, with the largest drops under Qwen3.6-27B.
Under Gemini, LLM-judge repair slightly exceeds our method in recovery, 0.740 versus 0.733, corresponding to only one additional recovered case.
However, our method has lower recurrence, 0.091 versus 0.108, perfect benign preservation, and lower replay and LLM cost. The lower benign preservation of LLM-judge repair under Gemini and Qwen reflects occasional over-repair, where the model removes or quarantines non-target memories. Overall, these supplemental results support the main experimental conclusions: backbone choice affects absolute recovery and repair-plan quality, but dependency-guided rollback continues to provide a stronger balance of recovery, recurrence, preservation, and cost.

\paragraph{Repair cost breakdown.}
Table~\ref{tab:app_cost_breakdown} shows that our method is not simply buying recovery with more computation. Compared with LLM-judge repair, our method improves recovery from 77.3\% to 85.3\% while using less than half the total tokens and fewer LLM calls. AgentTrace-style is cheaper, with a lower replay ratio of 9.1\%, but its recovery drops to 60.7\%, showing the cost of under-repairing persistent memory contamination. The ablations clarify the main cost driver. Removing selective replay raises the replay ratio from 12.3\% to 75.5\%, while recovery does not improve over the full method. Removing support checking mainly affects selectivity, and removing the rollback planner hurts both recovery and efficiency.

\begin{table*}[t!]
\centering
\small
\setlength{\tabcolsep}{4pt}
\begin{tabularx}{\textwidth}{l *{4}{>{\centering\arraybackslash}X}}
\hline
Method & Recovery $\uparrow$ & Total tokens $\downarrow$ & LLM count $\downarrow$ & Replay ratio $\downarrow$ \\
\hline
No repair & 0.000 & 0.00 & 0.00 & 0.000 \\
Full memory reset & 0.407 & 6509.33 & 4.03 & 0.061 \\
Delete retrieved memories & 0.340 & 7399.60 & 4.05 & 0.061 \\
MemAudit-style & 0.320 & 19700.95 & 10.43 & 0.051 \\
\hline
LLM-judge repair & 0.773 & 28175.28 & 9.80 & 0.217 \\
AgentTrace-style & 0.607 & 10508.94 & 4.69 & 0.091 \\
Ours & 0.853 & 12697.55 & 5.70 & 0.123 \\
\hline
w/o support check & 0.880 & 12954.37 & 6.51 & 0.154 \\
w/o rollback planner & 0.713 & 14969.29 & 9.19 & 0.152 \\
w/o selective replay & 0.840 & 55018.01 & 24.01 & 0.755 \\
\hline
\end{tabularx}
\caption{\textbf{Repair cost breakdown.} Our method achieves higher recovery than LLM-judge repair while using fewer tokens and fewer LLM calls. The ablations show that selective replay is the main mechanism preventing repair cost from increasing sharply.}
\label{tab:app_cost_breakdown}
\end{table*}

\paragraph{Sensitivity to fault and memory scale.}
Table~\ref{tab:app_sensitivity} reports robustness slices by faulty memory count and benign active memory store size. These are not controlled scaling laws because bucket sizes and fault compositions vary, but the method does not collapse on the available multi-fault cases. Recovery is 85.2\% for one-fault cases and 82.4\% for two-fault cases, while the three-fault bucket remains recoverable but contains only five cases. Memory store size is also non-monotonic. Recovery drops to 77.8\% in the 6--8 bucket but rises to 91.4\% in the 9--11 bucket. Replay ratio remains modest across buckets, ranging from 10.8\% to 15.5\%. These trends suggest that repair difficulty depends more on dependency shape, fault semantics, and whether the faulty memory lies on a load-bearing answer path than on memory store size alone.

\begin{table*}[t!]
\centering
\small
\setlength{\tabcolsep}{4pt}
\resizebox{\textwidth}{!}{%
\begin{tabular}{llrrrrrrr}
\hline
Slice & Bucket & $n$ & Recovery $\uparrow$ & Recurrence $\downarrow$ & Claim-inv.\ F1 $\uparrow$ & Replay ratio $\downarrow$ & Total tokens $\downarrow$ & LLM count $\downarrow$ \\
\hline
Faulty memories & 1 & 128 & 0.852 & 0.266 & 0.553 & 0.119 & 12559.85 & 5.68 \\
Faulty memories & 2 & 17 & 0.824 & 0.357 & 0.622 & 0.135 & 12586.29 & 5.59 \\
Faulty memories & 3 & 5 & 1.000 & 0.000 & 0.714 & 0.155 & 16601.00 & 6.60 \\
\hline
Memory store size & 0--5 & 27 & 0.889 & 0.583 & 0.617 & 0.149 & 8806.26 & 5.37 \\
Memory store size & 6--8 & 36 & 0.778 & 0.071 & 0.435 & 0.131 & 15304.03 & 6.50 \\
Memory store size & 9--11 & 58 & 0.914 & 0.321 & 0.575 & 0.110 & 12465.07 & 5.69 \\
Memory store size & $\geq$12 & 29 & 0.793 & 0.043 & 0.667 & 0.108 & 13549.83 & 5.03 \\
\hline
\end{tabular}%
}
\caption{\textbf{Robustness slices by faulty memory count and memory store size.} Memory store size is measured by the number of benign active memories. These slices are intended as robustness checks rather than scaling claims, since bucket sizes and fault compositions are not controlled independently.}
\label{tab:app_sensitivity}
\end{table*}

%% file: appendix/case_study.tex
\section{Case Study}
\label{app:case-study}

We present one running example, two compact cases from the controlled benchmark, and one multi-fault case from the adapted LongMemEval-V2 subset. The customer support example illustrates the full repair pipeline, while the travel and shopping cases highlight two complementary mechanisms: cleaning a wrong-user root memory together with contaminated derived memories, and replaying a load-bearing tool-call path after a poisoned factual memory. The final case illustrates how multiple poisoned memories can reinforce the same error across different levels of abstraction, requiring both persistent memory cleanup and trace-level replay for complete repair.

\subsection{Running Example: Customer Support Summary-Drift Case}

We use \texttt{customer\_support\_task\_036} because the fault propagates beyond the final answer into downstream claims and memory mutations.

\paragraph{Task context.}
\begin{quote}
\small
\textbf{User input:} Using only the latest concise support case summary you stored, tell me the order status and refund window together.

\textbf{Faulty answer:} The order status for \texttt{ORD00169} is shipped, and the refund window is \texttt{45 days}.

\textbf{Expected repaired answer:} Order \texttt{ORD00169} is shipped, and the Home refund window is \texttt{30 days}.
\end{quote}

\paragraph{Fault provenance.}
The failure is caused by a drifted memory update. The prior summary \texttt{m\_019} correctly records that order \texttt{ORD00169} is shipped and has a Home refund window of \texttt{30 days}. The faulty updated summary \texttt{m\_f026} changes the refund window to \texttt{45 days}. Given diagnosed faulty memory \texttt{m\_f026}, the provenance module follows the incoming \textsc{produce} edge to update step \texttt{s\_132}, which consumed \texttt{m\_019}, invalidated it, and produced the drifted summary. This identifies the fault as summary-drift during memory update rather than an isolated root memory store fault.

\paragraph{Affected state and support checking.}
Forward tracing from \texttt{s\_132} and \texttt{m\_f026} identifies 21 affected nodes: 5 memory nodes, 4 memory read steps, 4 claim steps, 4 final answer steps, 3 memory write steps, and 1 memory update step. The answer-relevant region contains the faulty final answer \texttt{s\_148}, which depends on a downstream claim using the drifted refund window.

Support checking prevents all reachable nodes from being treated as unsafe. After removing the faulty summary, the checker marks \texttt{s\_132} as independently supported, while downstream affected claims and answer steps remain unsupported. It also separates affected but supported derived memories from unsafe ones, preserving \texttt{m\_027}, \texttt{m\_028}, \texttt{m\_029}, and \texttt{m\_030} because they remain independently supported despite being reachable from the faulty summary.

\paragraph{Rollback and replay.}
The planner deletes \texttt{m\_f026}, replays the faulty summary producer and the answer-relevant downstream computation, and records unsupported but not answer-relevant steps as suspicious, invalidating them without replay. This follows the precedence rule of Section~\ref{sec:rollback-planning}: \texttt{s\_132} is replayed because it produced the faulty summary, even though its source evidence is independently supported. Conversely, \texttt{s\_137} is invalidated and recorded as suspicious because it is unsupported but not answer-relevant.

After replay, the faulty summary is replaced by corrected memory \texttt{m\_n034}, which states that the Home refund window is \texttt{30 days}. Unrelated support memories are preserved, \texttt{m\_028} remains active, and affected side memories are either preserved or regenerated according to their support status.

\paragraph{Corrected output and baseline behavior.}
The repaired final answer states that order \texttt{ORD00169} is shipped and that the Home refund window is \texttt{30 days}. No repair propagates the drifted summary. MemAudit-style preserves the drifted summary because memory auditing alone does not force replay of the affected summary and answer path. Delete retrieved memories removes useful context and yields an incomplete answer, while full reset deletes the support history needed to answer from memory. LLM-judge repair and AgentTrace-style recover the answer, but do not produce the same selective rollback account over the faulty producer and downstream state.

\subsection{Derived Faulty Memories: Travel Wrong-User Memory Case}

Instance \texttt{travel\_task\_048} asks for the airline of the exact confirmed active flight. The faulty answer is Qantas, while the expected repaired answer is ANA. The source fault is a wrong-user memory \texttt{m\_f000} stating that the active flight airline was corrected to Qantas rather than ANA. Since this memory has no producing step in the current trace, our method treats it as a root memory store fault.

The wrong-user memory also contaminates a downstream active flight summary. In the clean store, \texttt{m\_008} records flight \texttt{F0028} with airline ANA and price \texttt{\$275.99}. In the faulty run, the derived summary is rewritten with Qantas. Thus the final answer is supported both by the wrong-user root memory and by a contaminated derived memory. Our method deletes \texttt{m\_f000}, remaps the contaminated active flight summary to corrected memory \texttt{m\_n012}, and recovers ANA while preserving unrelated travel memories.

This case illustrates why persistent memory state repair is necessary. No repair and AgentTrace-style leave the root wrong-user memory and derived summary active, MemAudit-style removes the derived summary but misses the root memory, and LLM-judge repair fixes the immediate answer but misses the contaminated derived memory, leading to recurrence. Full reset and delete retrieved memories can recover the answer, but only by sacrificing substantially more benign state. Delete retrieved memories preserves only 6 in 10 benign memories, while full reset deletes all 12 active memories.

\subsection{Tool Call Replay: Shopping Poisoned Memory Case}

Instance \texttt{shopping\_task\_014} asks the agent to compare current prices for the remembered product. The faulty answer selects StoreB at \texttt{\$66.95} and the expected repaired answer selects StoreC at \texttt{\$63.99}. The source fault \texttt{m\_f003} poisons the StoreC price for product \texttt{P0199}, raising it to \texttt{\$88.67}. This makes the faulty answer locally consistent with corrupted price evidence, since StoreB appears cheaper than StoreC.

Reliable repair therefore requires more than deleting the poisoned memory or editing the final response. The load-bearing comparison path must be replayed so that \texttt{compare\_price(P0199)} refreshes the tool-derived evidence. Our method deletes \texttt{m\_f003}, replays the answer-relevant path including the comparison tool call, regenerates the price comparison memory, and answers StoreC at \texttt{\$63.99}. The replayed tool result reports StoreC at \texttt{\$63.99}, StoreA at \texttt{\$67.00}, and StoreB at \texttt{\$66.95}, which flips the final comparison while preserving other active product memories.

All baselines fail this case. Memory-centric methods either leave the poisoned price active or remove too much product context. LLM-judge repair deletes the poisoned memory, and AgentTrace-style avoids keeping it active, but neither recovers the corrected tool-derived evidence needed to change the final comparison. The case shows why dependency-guided rollback must sometimes replay the tool action that produced or refreshed contaminated evidence.

\subsection{Multi-Fault: Adapted LongMemEval-V2 Case}

We include one adapted LongMemEval-V2 case to illustrate how multiple poisoned memories can jointly contaminate execution. Instance \texttt{lme\_task\_025\_901b7d17} asks the agent to identify the new header action button that appears after opening a ServiceNow knowledge article from the search-results page. The expected answer is \texttt{Attach to Private Task}, while the faulty execution answers \texttt{Add to Private Task}.

This case contains three poisoned memories that encode the same wrong conclusion at different abstraction levels. Memory \texttt{m\_f005} corrupts the retrieved evidence state by recording the article-page header button as \texttt{Add to Private Task}. Memory \texttt{m\_f006} turns this corrupted observation into a derived factual statement, claiming that \texttt{Add to Private Task} is the label visible near the article heading and absent from the search-results context. Memory \texttt{m\_f009} further consolidates the error into a compact answer memory. The new article-page header action label is \texttt{Add to Private Task}. Thus the final answer is not caused by a single isolated poisoned fact. The same wrong label is reinforced across raw evidence, extracted observation, and compact answer memory.

Starting from the diagnosed faulty memories, our method deletes \texttt{m\_f005}, \texttt{m\_f006}, and \texttt{m\_f009}. During rollback, it invalidates the affected claim steps \texttt{s\_046} and \texttt{s\_049}, replays the answer-relevant path, and reruns the evidence lookup needed to refresh the article-page observation. The repaired execution generates replacement memories \texttt{m\_n010}, \texttt{m\_n011}, and \texttt{m\_n012}. These replacements consistently record the corrected label \texttt{Attach to Private Task}: the repaired raw evidence memory stores the corrected article-page button, the repaired derived observation states that this label appears on the article page, and the repaired compact answer memory identifies it as the new header action. The corrected final answer is therefore \texttt{Attach to Private Task}.

This case highlights why multi-level poisoned memories require both memory-state cleanup and trace-level replay. No repair keeps the poisoned memories active and repeats the faulty answer. Memory-centric baselines delete the poisoned memories but fail to recover the answer. MemAudit-style answers \texttt{Subscribe}, while full reset and delete retrieved memories answer \texttt{Edit} after losing useful context. LLM-judge repair also deletes all three poisoned memories but still regenerates \texttt{Add to Private Task}. AgentTrace-style recovers the final answer through trace-level replay, but without the same explicit cleanup account over all jointly poisoned memory levels. Our method recovers the answer while deleting all three poisoned memories, preserving all benign memories, and regenerating consistent replacement memories for the raw evidence, derived observation, and compact answer.

%% file: appendix/baseline.tex
\section{Baseline Implementation Details}
\label{app:baseline}

We summarize the baseline implementations used in our evaluation. Since prior methods do not directly target post-failure repair for memory-augmented executions, we adapt each method to the same repair interface. Each method receives the failed trace, faulty memory store, diagnosed faulty memory ids, and final turn task input, then outputs either a memory edit decision or a rollback plan executed by the shared replay executor. Table~\ref{tab:baseline_comparison} summarizes the configurations.

\begin{table*}[t]
\centering
\small
\setlength{\tabcolsep}{5pt}
\begin{tabular}{llccc}
\hline
Method & Repair level & Memory repair & Trace rollback & Dependency graph \\
\hline
No repair & None & No & No & No \\
Full memory reset & Memory & All memories & No & No \\
Delete retrieved memories & Memory & Retrieved memories & No & No \\
MemAudit-style & Memory & Ranked memories & No & Memory only \\
LLM-judge repair & Trace & LLM-selected & LLM-selected & No \\
AgentTrace-style & Trace & Indirect & Suspicious steps & Trace only \\
Ours & Memory+trace & Selective & Selective & Memory+trace \\
\hline
\end{tabular}
\caption{\textbf{Comparison of repair baselines.} Baselines differ in whether they repair memory state, roll back faulty trace state, or use dependency structure.}
\label{tab:baseline_comparison}
\end{table*}

\paragraph{No Repair.}
This lower-bound condition keeps the failed run unchanged, including the faulty memory store, faulty trace, and original final answer. It emits an empty repair plan and performs no cleanup, deletion, rollback, or replay.

\paragraph{Full Memory Reset.}
This memory-centric baseline removes all active memories before replaying the final user turn under the same agent lifecycle. The trace prefix before the final turn is kept fixed, and the regenerated final turn trace is normalized to the same artifact schema as other methods. This strategy removes all potentially contaminated memory, but also deletes benign personalization state.

\paragraph{Delete Retrieved Memories.}
This baseline deletes active memories that appear in the affected region of the failed execution before the final turn. Starting from the affected subgraph seeded by the diagnosed faulty memories, it collects active memories retrieved or referenced by relevant non-retrieval steps, where non-retrieval steps refer to all steps except memory read steps, then deletes those memories and replays the final user turn. It does not perform independent-support checking, derived memory cleanup, or answer-relevant rollback optimization.

\paragraph{LLM-Judge Repair.}
This baseline asks an LLM to predict a rollback plan from a compact view of the failed execution, including session turns, diagnosed faulty memory ids, compact memory records, compact trace steps, the wrong final answer, valid id sets, and the executor output schema. The output is constrained to a JSON rollback plan with fields for memory deletion, memory quarantine, replay steps, preserved steps, and suspicious steps. The plan must reference only provided memory and step identifiers and is required to delete every diagnosed faulty memory.

We validate each generated plan before execution. Plans are rejected if they contain invalid JSON, malformed fields, invalid ids, overlapping step dispositions, or missing deletion of diagnosed faulty memories. Validation failures are retried with the error appended to the prompt. If no valid plan is produced within the retry budget, the case is marked as failed. Valid plans are executed by the same rollback executor as our method, isolating the effect of the planning strategy.

\paragraph{MemAudit-style.}
We adapt MemAudit-style post-hoc memory auditing~\cite{tan2026memaudit} as a memory deletion selector. The candidate pool contains active memories before the final user turn that are used by downstream non-retrieval steps. If this pool is empty, we fall back to diagnosed source fault memories that remain active. We report this fallback as an oracle-assisted diagnostic fallback used only to avoid an empty audit candidate set, not as an unsupervised MemAudit signal. Candidate memories are scored using a memory consistency anomaly signal and a counterfactual influence signal. The consistency graph compares memories with matching entity and fact fields. Matching values induce support edges, while different values induce contradiction edges.

For candidate memory \(m\), the anomaly score is
\[
\mathrm{anomaly}(m)=
\frac{\mathrm{contradiction}(m)}
{\mathrm{support}(m)+\mathrm{contradiction}(m)},
\]
with score \(0\) when the denominator is zero. In the oracle-assisted diagnostic variant, if benchmark provided faulty content directly matches a memory, that memory is assigned anomaly score \(1.0\). Counterfactual influence is computed by deleting \(m\), rerunning the final user turn, and scoring the regenerated trace with the deterministic task oracle:
\[
\mathrm{influence}(m)=
\frac{\max(0, S_{-m}-S_{\mathrm{orig}})}
{1-S_{\mathrm{orig}}},
\]
where \(S_{\mathrm{orig}}\) is the original faulty trace score and \(S_{-m}\) is the score after deleting \(m\). If the original score is already \(1\), or if the rerun fails, the influence score is set to \(0\). The final score is
\[
\mathrm{score}(m)=
\alpha \cdot \mathrm{influence}(m)
+(1-\alpha)\cdot \mathrm{anomaly}(m),
\]
with \(\alpha=0.5\). Memories with score above \(0.5\) are deleted together, and only the final user turn is replayed. This baseline does not roll back trace nodes, invalidate claims, repair derived memory lineage, or select answer relevant replay frontiers.

\paragraph{AgentTrace-style.}
We adapt AgentTrace-style causal trace localization~\cite{wang2026agenttrace} as a trace-centric repair baseline. The method receives the same diagnosed faulty memory ids as other methods, constructs faulty memory provenance and affected regions, scores candidate trace steps, selects the highest ranked step as the root cause, and converts it into a replay plan executed by the shared rollback executor. We use this adaptation only to choose replay regions, not to implement a full persistent memory repair workflow.

Candidate steps are restricted to memory read, claim, plan, tool action, and final answer steps that are backward reachable from the erroneous final answer and fault-relevant. A step is fault-relevant if it lies in the affected trace region, directly uses faulty memory, belongs to the fault dependent suffix, or is the final answer step. Each candidate \(v\) receives
\[
\begin{aligned}
r(v)=&
0.40 f_{\mathrm{back}}(v)
+0.25 f_{\mathrm{aff}}(v)
+0.20 f_{\mathrm{fault}}(v) \\
&+0.10 f_{\mathrm{down}}(v)
+0.05 f_{\mathrm{type}}(v),
\end{aligned}
\]
where \(f_{\mathrm{back}}=1-d/d_{\max}\) measures backward relevance, \(f_{\mathrm{aff}}\) and \(f_{\mathrm{fault}}\) indicate affected region membership and direct faulty memory involvement, \(f_{\mathrm{down}}=\min(1,\mathrm{reached\_steps}/8)\) measures downstream reachability and is raised to at least \(0.5\) if the step reaches the erroneous final answer, and \(f_{\mathrm{type}}\) is a fixed step type prior set to \(0.95\), \(0.85\), \(0.75\), \(0.65\), and \(0.40\) for memory read, claim, plan, tool action, and final answer steps, respectively. The weights and priors are fixed across all domains and fault types.

If no valid candidate is found, the baseline falls back to the final answer step. The replay region combines provenance source steps for diagnosed faulty memories with the sequential suffix from the selected root cause step through the erroneous final answer. Persistent memory cleanup is disabled, so existing faulty memories remain active unless overwritten as a side effect of replay.

%% file: appendix/adapted_lmev2.tex
\section{Adapted LongMemEval-V2 Subset Details}
\label{app:adapted-lmev2}

\paragraph{Subset Construction.}
We adapt a subset of LongMemEval-V2~\cite{wu2026longmemeval}, which consists of long-context user-assistant interaction trajectories. We do not use the full benchmark distribution. Instead, we retain only cases for which our pipeline can construct reliable clean data for memory repair evaluation. Specifically, the trajectory must contain sufficient textual evidence to recover the clean answer, the evidence must be convertible into our controlled benchmark schema, and the clean answer must be verifiable without relying on hidden gold labels.

This selection criterion favors procedural and navigation-style tasks, whose answers are grounded in explicit trajectory evidence such as page titles, menu labels, form fields, or ordered action sequences. We exclude cases that cannot be reliably converted into clean instances, including ambiguous boolean questions, generic short-answer questions, visual-only UI states, and aggregation questions whose required evidence is not explicit in the retained trajectory text. The resulting adapted set contains 50 cases.

\paragraph{Schema Adaptation.}
Each selected trajectory is converted into the same schema used by our controlled benchmark. Trajectory observations, action traces, page states, and supporting evidence snippets are mapped into memory records with provenance metadata. Fault construction follows the poisoned memory procedure described in Appendix~\ref{app:fault-injection}.

Final answers are evaluated with the deterministic LongMemEval-V2-derived oracle described in Appendix~\ref{app:metrics}. When an answer contains multiple fields or ordered steps, phrase-set matching checks each required element, and ordered matching is applied when the original evaluator requires order.

\paragraph{Fault Distribution.}
Table~\ref{tab:lmev2-fault-distribution} shows the number of faulty memories per adapted case. Most cases in this subset contain multiple faulty memories, complementing our controlled benchmark, where most cases contain a single faulty memory.

\begin{table}[h]
\centering
\small
\begin{tabular}{lc}
\toprule
Number of faulty memories & Number of cases \\
\midrule
1 & 5 \\
2 & 5 \\
3 & 38 \\
4 & 2 \\
\bottomrule
\end{tabular}
\caption{Fault multiplicity in the adapted LongMemEval-V2 subset.}
\label{tab:lmev2-fault-distribution}
\end{table}

\paragraph{Limitations.}
The adapted LongMemEval-V2 subset is intended as a transfer stress test rather than a replacement for the full LongMemEval-V2 benchmark. Because we retain only cases from which reliable clean data can be constructed under our repair schema, the subset does not represent the full LongMemEval-V2 distribution and should not be used to claim general LongMemEval-V2 performance. Its purpose is to test whether repair behavior transfers from our controlled benchmark to trajectory-derived memory records.

%% file: references.bib
@inproceedings{park2023generative,
  title={Generative agents: Interactive simulacra of human behavior},
  author={Park, Joon Sung and O'Brien, Joseph and Cai, Carrie Jun and Morris, Meredith Ringel and Liang, Percy and Bernstein, Michael S},
  booktitle={Proceedings of the 36th annual acm symposium on user interface software and technology},
  pages={1--22},
  year={2023}
}

@article{packer2023memgpt,
  title={MemGPT: towards LLMs as operating systems.},
  author={Packer, Charles and Fang, Vivian and Patil, Shishir G. and Lin, Kevin and Wooders, Sarah and Gonzalez, Joseph E.},
  year={2023},
  publisher={ArXiv}
}

@article{zhang2025survey,
  title={A survey on the memory mechanism of large language model-based agents},
  author={Zhang, Zeyu and Dai, Quanyu and Bo, Xiaohe and Ma, Chen and Li, Rui and Chen, Xu and Zhu, Jieming and Dong, Zhenhua and Wen, Ji-Rong},
  journal={ACM Transactions on Information Systems},
  volume={43},
  number={6},
  pages={1--47},
  year={2025},
  publisher={ACM New York, NY}
}

@inproceedings{zhong2024memorybank,
  title={Memorybank: Enhancing large language models with long-term memory},
  author={Zhong, Wanjun and Guo, Lianghong and Gao, Qiqi and Ye, He and Wang, Yanlin},
  booktitle={Proceedings of the AAAI conference on artificial intelligence},
  volume={38},
  number={17},
  pages={19724--19731},
  year={2024}
}

@article{wang2023voyager,
  title={Voyager: An open-ended embodied agent with large language models},
  author={Wang, Guanzhi and Xie, Yuqi and Jiang, Yunfan and Mandlekar, Ajay and Xiao, Chaowei and Zhu, Yuke and Fan, Linxi and Anandkumar, Anima},
  journal={arXiv preprint arXiv:2305.16291},
  year={2023}
}

@inproceedings{xiong2026memory,
  title={How memory management impacts llm agents: An empirical study of experience-following behavior},
  author={Xiong, Zidi and Lin, Yuping and Xie, Wenya and He, Pengfei and Liu, Zirui and Tang, Jiliang and Lakkaraju, Himabindu and Xiang, Zhen},
  booktitle={Proceedings of the 64th Annual Meeting of the Association for Computational Linguistics (Volume 1: Long Papers)},
  pages={623--645},
  year={2026}
}

@article{chao2026stale,
  title={STALE: Can LLM Agents Know When Their Memories Are No Longer Valid?},
  author={Chao, Hanxiang and Bai, Yihan and Sheng, Rui and Li, Tianle and Sun, Yushi},
  journal={arXiv preprint arXiv:2605.06527},
  year={2026}
}

@article{chen2025halumem,
  title={Halumem: Evaluating hallucinations in memory systems of agents},
  author={Chen, Ding and Niu, Simin and Li, Kehang and Liu, Peng and Zheng, Xiangping and Tang, Bo and Li, Xinchi and Xiong, Feiyu and Li, Zhiyu},
  journal={arXiv preprint arXiv:2511.03506},
  year={2025}
}

@article{shinn2023reflexion,
  title={Reflexion: Language agents with verbal reinforcement learning},
  author={Shinn, Noah and Cassano, Federico and Gopinath, Ashwin and Narasimhan, Karthik and Yao, Shunyu},
  journal={Advances in neural information processing systems},
  volume={36},
  pages={8634--8652},
  year={2023}
}

@article{tan2026memaudit,
  title={MemAudit: Post-hoc Auditing of Poisoned Agent Memory via Causal Attribution and Structural Anomaly Detection},
  author={Tan, Zhewen and Yao, Yilun and Jin, Huiyan and Yu, Wenhan and Wang, Guoan and Fan, Mengyuan and Liu, Feng and Zhang, Xiangzheng and Ma, Duohe and Yang, Tong and others},
  journal={arXiv preprint arXiv:2605.23723},
  year={2026}
}

@inproceedings{zou2025poisonedrag,
  title={$\{$PoisonedRAG$\}$: Knowledge corruption attacks to $\{$Retrieval-Augmented$\}$ generation of large language models},
  author={Zou, Wei and Geng, Runpeng and Wang, Binghui and Jia, Jinyuan},
  booktitle={34th USENIX Security Symposium (USENIX Security 25)},
  pages={3827--3844},
  year={2025}
}

@article{xu2026mem,
  title={A-mem: Agentic memory for llm agents},
  author={Xu, Wujiang and Liang, Zujie and Mei, Kai and Gao, Hang and Tan, Juntao and Zhang, Yongfeng},
  journal={Advances in Neural Information Processing Systems},
  volume={38},
  pages={17577--17604},
  year={2026}
}

@article{zhao2026ama,
  title={AMA-Bench: Evaluating long-horizon memory for agentic applications},
  author={Zhao, Yujie and Yuan, Boqin and Huang, Junbo and Yuan, Haocheng and Yu, Zhongming and Xu, Haozhou and Hu, Lanxiang and Shankarampeta, Abhilash and Huang, Zimeng and Ni, Wentao and others},
  journal={arXiv preprint arXiv:2602.22769},
  year={2026}
}

@inproceedings{du2026memguide,
  title={MemGuide: intent-driven memory selection for goal-oriented multi-session LLM agents},
  author={Du, Yiming and Wang, Bingbing and He, Yang and Liang, Bin and Wang, Baojun and Li, Zhongyang and Gui, Lin and Pan, Jeff Z and Xu, Ruifeng and Wong, Kam-Fai},
  booktitle={Proceedings of the AAAI Conference on Artificial Intelligence},
  volume={40},
  number={36},
  pages={30584--30592},
  year={2026}
}

@article{yao2022react,
  title={React: Synergizing reasoning and acting in language models},
  author={Yao, Shunyu and Zhao, Jeffrey and Yu, Dian and Du, Nan and Shafran, Izhak and Narasimhan, Karthik and Cao, Yuan},
  journal={arXiv preprint arXiv:2210.03629},
  year={2022}
}

@article{yao2024tau,
  title={$\tau$-Bench: A Benchmark for Tool-Agent-User Interaction in Real-World Domains},
  author={Yao, Shunyu and Shinn, Noah and Razavi, Pedram and Narasimhan, Karthik},
  journal={arXiv preprint arXiv:2406.12045},
  year={2024}
}

@article{deng2023mind2web,
  title={Mind2web: Towards a generalist agent for the web},
  author={Deng, Xiang and Gu, Yu and Zheng, Boyuan and Chen, Shijie and Stevens, Sam and Wang, Boshi and Sun, Huan and Su, Yu},
  journal={Advances in Neural Information Processing Systems},
  volume={36},
  pages={28091--28114},
  year={2023}
}

@article{xue2025illusion,
  title={An illusion of progress? assessing the current state of web agents},
  author={Xue, Tianci and Qi, Weijian and Shi, Tianneng and Song, Chan Hee and Gou, Boyu and Song, Dawn and Sun, Huan and Su, Yu},
  journal={arXiv preprint arXiv:2504.01382},
  year={2025}
}

@article{sunil2026memory,
  title={Memory poisoning attack and defense on memory based LLM-agents},
  author={Sunil, Balachandra Devarangadi and Sinha, Isheeta and Maheshwari, Piyush and Todmal, Shantanu and Mallik, Shreyan and Mishra, Shuchi},
  journal={arXiv preprint arXiv:2601.05504},
  year={2026}
}

@inproceedings{hatalis2023memory,
  title={Memory matters: The need to improve long-term memory in llm-agents},
  author={Hatalis, Kostas and Christou, Despina and Myers, Joshua and Jones, Steven and Lambert, Keith and Amos-Binks, Adam and Dannenhauer, Zohreh and Dannenhauer, Dustin},
  booktitle={Proceedings of the AAAI Symposium Series},
  volume={2},
  number={1},
  pages={277--280},
  year={2023}
}

@article{zhang2026useful,
  title={Useful memories become faulty when continuously updated by llms},
  author={Zhang, Dylan and Lin, Yanshan and Wu, Zhengkun and Sun, Yihang and Li, Bingxuan and Li, Dianqi and Peng, Hao},
  journal={arXiv preprint arXiv:2605.12978},
  year={2026}
}

@article{chen2024agentpoison,
  title={Agentpoison: Red-teaming llm agents via poisoning memory or knowledge bases},
  author={Chen, Zhaorun and Xiang, Zhen and Xiao, Chaowei and Song, Dawn and Li, Bo},
  journal={Advances in Neural Information Processing Systems},
  volume={37},
  pages={130185--130213},
  year={2024}
}

@article{madaan2023self,
  title={Self-refine: Iterative refinement with self-feedback},
  author={Madaan, Aman and Tandon, Niket and Gupta, Prakhar and Hallinan, Skyler and Gao, Luyu and Wiegreffe, Sarah and Alon, Uri and Dziri, Nouha and Prabhumoye, Shrimai and Yang, Yiming and others},
  journal={Advances in neural information processing systems},
  volume={36},
  pages={46534--46594},
  year={2023}
}

@article{wang2026agenttrace,
  title={AgentTrace: Causal Graph Tracing for Root Cause Analysis in Deployed Multi-Agent Systems},
  author={Wang, Zhaohui Geoffrey},
  journal={arXiv preprint arXiv:2603.14688},
  year={2026}
}

@article{wu2026longmemeval,
  title={Longmemeval-v2: Evaluating long-term agent memory toward experienced colleagues},
  author={Wu, Di and Ji, Zixiang and Kawatkar, Asmi and Kwan, Bryan and Gu, Jia-Chen and Peng, Nanyun and Chang, Kai-Wei},
  journal={arXiv preprint arXiv:2605.12493},
  year={2026}
}

@article{zhao2026memorepair,
  title={{MEMOREPAIR}: Barrier-First Cascade Repair in Agentic Memory},
  author={Zhao, Yang and Dai, Chengxiao and Kou, Mengying and Xiu, Yue},
  journal={arXiv preprint arXiv:2605.07242},
  year={2026}
}

@article{weiser1984program,
  title={Program Slicing},
  author={Weiser, Mark},
  journal={IEEE Transactions on Software Engineering},
  volume={SE-10},
  number={4},
  pages={352--357},
  year={1984},
  doi={10.1109/TSE.1984.5010248}
}

@inproceedings{agrawal1990dynamic,
  title={Dynamic Program Slicing},
  author={Agrawal, Hiralal and Horgan, Joseph R.},
  booktitle={Proceedings of the ACM SIGPLAN 1990 Conference on Programming Language Design and Implementation},
  pages={246--256},
  year={1990},
  doi={10.1145/93542.93576}
}

@article{cheney2009provenance,
  title={Provenance in Databases: Why, How, and Where},
  author={Cheney, James and Chiticariu, Laura and Tan, Wang-Chiew},
  journal={Foundations and Trends in Databases},
  volume={1},
  number={4},
  pages={379--474},
  year={2009},
  doi={10.1561/1900000006}
}

@article{mohan1992aries,
  title={{ARIES}: A Transaction Recovery Method Supporting Fine-Granularity Locking and Partial Rollbacks Using Write-Ahead Logging},
  author={Mohan, Chandrasekaran and Haderle, Don and Lindsay, Bruce and Pirahesh, Hamid and Schwarz, Peter},
  journal={ACM Transactions on Database Systems},
  volume={17},
  number={1},
  pages={94--162},
  year={1992},
  doi={10.1145/128765.128770}
}
